\documentclass{article}
\usepackage{arxiv}
\usepackage{amsmath,amsfonts}
\usepackage{algorithmic}
\usepackage{algorithm}
\usepackage{array}
\usepackage[caption=false,font=normalsize,labelfont=sf,textfont=sf]{subfig}
\usepackage{textcomp}
\usepackage{stfloats}
\usepackage{url}
\usepackage{verbatim}
\usepackage{graphicx}
\usepackage{cite}
\usepackage{booktabs}
\usepackage{multirow}
\usepackage{mathtools}
\begin{document}

\title{ReinVBC: A Model-based Reinforcement Learning Approach to Vehicle Braking Controller}
\author{%
    Haoxin Lin\textsuperscript{\rm 1,\rm 2,\rm 3},~~
    Junjie Zhou\textsuperscript{\rm 3 $\ast$},~~
    Daheng Xu\textsuperscript{\rm 3}\thanks{Work performed while Junjie Zhou and Daheng Xu were at Polixir Technologies},~~
    Yang Yu\textsuperscript{\rm 1,\rm 2,\rm 3}\thanks{Corresponding Author} \\   
    \textsuperscript{\rm 1}National Key Laboratory for Novel Software Technology, Nanjing University, Nanjing, China\\
    \textsuperscript{\rm 2}School of Artificial Intelligence, Nanjing University, Nanjing, China\\
    \textsuperscript{\rm 3}Polixir Technologies, Nanjing, China\\
    linhx@lamda.nju.edu.cn, yuy@nju.edu.cn
}
\date{}



\maketitle

\begin{abstract}
Braking system, the key module to ensure the safety and steer-ability of current vehicles, relies on extensive manual calibration during production. Reducing labor and time consumption while maintaining the Vehicle Braking Controller (VBC) performance greatly benefits the vehicle industry. Model-based methods in offline reinforcement learning, which facilitate policy exploration within a data-driven dynamics model, offer a promising solution for addressing real-world control tasks. This work proposes ReinVBC, which applies an offline model-based reinforcement learning approach to deal with the vehicle braking control problem. We introduce useful engineering designs into the paradigm of model learning and utilization to obtain a reliable vehicle dynamics model and a capable braking policy. Several results demonstrate the capability of our method in real-world vehicle braking and its potential to replace the production-grade anti-lock braking system.
\end{abstract}

\section{Introduction}
The braking system \cite{brake} is a critical module for vehicle chassis motion control. When a vehicle performs emergency braking under extreme road conditions, the braking system can control the cylinder pressure of all four wheels to prevent wheel lock-up, ensuring the vehicle's safety and steer-ability. Currently, most production-grade vehicles apply the Anti-lock Braking System (ABS) \cite{abs1,abs2,abs3,abs4} designed by Bosch, which performs well in terms of braking distance and braking deviation in most scenarios. However, ABS is implemented based on traditional controllers, and its parameters rely on extensive manual calibration in various scenarios. Moreover, the manually tuned parameters may not result in an optimal control. Searching for a nearly optimal policy that can adapt to most braking scenarios automatically without depending on manual experience is expected.

Reinforcement Learning (RL) \cite{rl,dqn,sac} is a promising way to find optimal control in sequential decision problems. Unfortunately, the success of RL has been primarily limited to simulators. The reason is that RL requires a lot of trial and error in the environment, which is unacceptable in the real world. Trial-and-error exploration not only consumes a considerable quantity of time and hardware resources but can also lead to dangerous accidents, especially in scenarios of robotic control \cite{robot1,robot2}. Several works \cite{rl_brake1,rl_brake2,rl_brake3,rl_brake4,rl_brake5,rl_brake6,rl_brake7} applying RL to vehicle braking remain in physical simulators or hardware-in-loop simulations rather than real-world environments.

Model-based Reinforcement Learning (MBRL) \cite{mopo,morel,combo,mobile,morec} in the offline setting appears to be a reliable solution for addressing real-world control tasks. The offline setting only requires collecting some samples in the real environment, where data collection can be performed using existing policies that may have low performance but ensure safety. The offline dataset can supervise the learning of the dynamics model, in which many explorations and evaluations of the agent can happen, thereby reducing the reliance on real-world samples. This paradigm has already achieved a great performance on D4RL \cite{d4rl} and NeoRL \cite{neorl} benchmarks. The capability of offline MBRL in real-world applications deserves further study.

This paper applies offline MBRL to vehicle braking control. During data collection, the braking controller can use an easily designed rule-based policy in simple scenarios to ensure safety. Since the causal relationships in vehicle dynamics are relatively clear, a reliable vehicle dynamics model can be learned from the dataset after manually defining the causal graph. Then, any classical RL algorithm can find a nearly optimal braking policy within this dynamics model. In general, our contributions are summarized as follows.
\begin{itemize}
    \item We set up several basic braking scenarios and design a simple control rule to collect data at low vehicle speeds, ensuring safety and testing the generalization capability of the learned policy in more complex braking tasks with riskier road surfaces and higher braking speeds.
    \item We design the state space, action space, and reward function for the decision-making process during vehicle braking. Then we propose an effective approach called ReinVBC (Model-based \textbf{Rein}forcement Learning to \textbf{V}ehicle \textbf{B}raking \textbf{C}ontrol), which learns a vehicle dynamics model according to the predefined causal graph and optimizes the policy in the model. The results demonstrate the long-term predictive ability of the learned dynamics model and the learning progress of the policy.
    \item In-distribution test shows that the braking controller by offline MBRL, \emph{i.e.}, ReinVBC, can outperform the original-equipment ABS controller, the simple rule-based policy used for data collection, and the direct braking without any control in the real-world scenarios covered by the data.
    \item Out-of-distribution test shows that the learned policy can achieve a small braking deviation and avoid the wheel lock-up during braking in both the hardware-in-loop simulation of complicated conditions and the real-world complex scenarios of the professional automotive testing ground, ensuring the safety and steer-ability.
\end{itemize}

Although the braking controller by offline MBRL cannot outperform the production-grade ABS thoroughly at present, this paper has demonstrated the potential of offline MBRL in real-world vehicle braking. We expect to replace the manual calibration of the traditional controller with a reliable data-driven learning paradigm, reducing the cost of production.

\section{Preliminaries}

\subsection{Vehicle Variables}
We list the vehicle variables involved in this paper and their corresponding descriptions in Table \ref{var}. Static parameters are inherent to vehicle manufacturing and are considered constants. Operational variables are determined by the driver’s operations, while control variables come from the controller’s decisions. Together, operational variables and control variables determine the dynamic variables during the vehicle’s movement.

\begin{table*}[pt!]
    \centering
    \caption{Important vehicle variables.}
    \resizebox{\linewidth}{!}{
    \begin{tabular}{r|llll}
        \toprule
        \textbf{Type} & \textbf{Name} & \textbf{Notation} & \textbf{Unit} & \textbf{Description} \\

        \midrule
        \multirow{5}{*}{dynamic variable}
        & vehicle speed & $v$ & $\mathrm{km}/\mathrm{h}$ & speed in the vehicle direction \\
        & wheel speed & $\mathbf{w}=(w^1, w^2, w^3, w^4)$ & $\mathrm{km}/\mathrm{h}$ & linear speed of the four wheels rotating \\
        & acceleration & $\mathbf{a}=(a_x, a_y, a_z)$ & $\mathrm{m}/\mathrm{s}^2$ & acceleration in the $x$, $y$, $z$ directions \\
        & attitude rate & $\dot{\mathbf{\omega}}=(\dot{\theta}, \dot{\phi}, \dot{\psi})$& $\mathrm{rad}/\mathrm{s}$ & change rates of pitch, roll, and yaw angles\\
        & wheel cylinder pressure & $\mathbf{p}=(p^1,p^2,p^3,p^4)$ & $\mathrm{MPa}$ & hydraulic pressure within the wheel cylinders \\
        \midrule
        control variable & wheel action & $\mathbf{u}=(u^1,u^2,u^3,u^4)$ & N/A & discrete instruction to control the wheel cylinder pressure \\
        \midrule
        \multirow{3}{*}{operational variable}
        & brake pedal force & $f_{\mathrm{brake}}$ & $\mathrm{N}$ & force applied to the vehicle's brake pedal \\
        & accelerator pedal force & $f_{\mathrm{acc}}$ & $\mathrm{N}$ & force applied to the vehicle's accelerator pedal \\
        & steering wheel angle & $\delta$ & $\mathrm{rad}$ & rotation angle of the steering wheel \\
        \midrule
        \multirow{2}{*}{static parameter}
        & wheelbase & $L_{\mathrm{veh}}$ & $\mathrm{m}$ & wheelbase of the vehicle \\
        & steering ratio & $N_s$ & N/A & ratio of the steering wheel angle to the steering angle\\
        \bottomrule
    \end{tabular}
    }
    \label{var}
\end{table*}

\subsection{Braking System}
Emergency braking is unavoidable for drivers in hazardous situations. When the driver presses the brake pedal heavily, the four wheels will tend to lock up if there is no additional pressure control. The locking condition of the $i$-th wheel is quantified by its slip ratio defined as $\eta^i =1-\frac{w^i}{v}$. A slip ratio of one indicates zero wheel speed, signifying wheel lock-up. Wheel lock-up can make the vehicle out of control, especially on extreme surfaces like icy roads. Therefore, the braking system \cite{brake} is designed to control the wheel cylinder pressure to maintain steer-ability and safety of the vehicle during heavy braking. The most widely used system for this purpose is the Anti-lock Braking System (ABS) \cite{abs1,abs2,abs3,abs4}, which aims to control the wheel slip ratio within a safe range.

\subsection{Markov Decision Process}
We consider a standard Markov decision process (MDP) specified by a tuple $\mathcal{M}=\langle \mathcal{S},\mathcal{U},T,r,\rho_0,\gamma \rangle$, where $\mathcal{S}$ is the state space, $\mathcal{U}$ is the action space, $T(s_{t+1}|s_t,u_t)$ is the dynamics function that calculates the conditioned distribution of $s_{t+1}\in\mathcal{S}$ given the state-action pair $(s_t,u_t)$, $r(s_t,u_t)$ is the reward function, $\rho_0$ is the initial state distribution, and $\gamma$ is the discount factor. 

Practical decision-making scenarios are generally not fully observable, as is the case with the brake control task discussed in this paper. Due to factors such as changing road conditions and sensor errors, the dynamic variables listed in Table \ref{var} cannot fully reflect the vehicle’s state. Their value at step t can only be considered as part of the state, denoted as the observation $o_t$. Typically, $o_t$ is stacked over a certain number of steps to approximate $s_t$:
\begin{equation}
s_t\approx (o_{t-h+1},\cdots,o_t),
\end{equation}
where $h$ is the number of stacked steps.

\subsection{Offline Model-based Reinforcement Learning}
We use $\rho_\pi$ to denote the on-policy distribution over states induced by the dynamics function $T$ and the policy $\pi$. The optimization goal of reinforcement learning \cite{rl} is to search a policy $\pi$ that maximized the expected discounted cumulative reward $\mathbb{E}_{\rho^\pi}\left[\sum_{t=0}^\infty \gamma^t r_t\right]$. Such a policy can be derived from the estimation of the state-action value function
$Q^\pi(s_t,u_t)=\mathbb{E}_{(s_{t+1},r_{t+1})\sim T(\cdot|s_t,u_t)}\left[r_{t+1}+\gamma V^\pi(s_{t+1})\right]$, where $V^\pi(s_{t+1})=\mathbb{E}_{u_{t+1}\sim\pi(\cdot|s_{t+1})}\left[Q^\pi(s_{t+1},u_{t+1})\right]$ is the state value function.

In the offline setting \cite{bcq, bear, cql, edac}, environmental samples are confined to a given static dataset $\mathcal{D}_{\mathrm{env}}$ since online exploration is inaccessible to the agent.
Offline MBRL \cite{mopo, morel, combo, mobile} aims to optimize the policy using the model augmented data beyond the dataset. The dynamics model $\hat{T}$ is typically trained to maximize the expected likelihood 
\begin{equation}
J(\hat{T})=\mathbb{E}_{(s_t,u_t,s_{t+1})\sim\mathcal{D}_{\mathrm{env}}}[\log\hat{T}(s_{t+1}|s_t,u_t)].
\end{equation} 
The estimated dynamics model defines a surrogate MDP $\hat{\mathcal{M}}=(\mathcal{S}, \mathcal{U}, \hat{T}, r, \rho_0, \gamma)$. 

The limited dataset causes $\hat{T}$ to cover only a part of the state-action space. Once the policy lead the trajectory to extend into out-of-distribution areas during roll-out in $\hat{\mathcal{M}}$, the learning process can collapse. Thus, MOPO \cite{mopo} and some of its subsequent offline MBRL algorithms \cite{morel, mobile} incorporate a penalty term measuring the model uncertainty in the reward function, allowing the agent to sample within safe regions of $\hat{T}$. Then any RL algorithm can be used to recover the optimal policy using the penalized reward function with the augmented dataset $\mathcal{D}_{\mathrm{env}}\cup\mathcal{D}_{\mathrm{model}}$, where $\mathcal{D}_{\mathrm{model}}$ is the synthetic data rolled out in $\hat{\mathcal{M}}$.

Nevertheless, this paper does not intend to introduce a penalty based on the model uncertainty into the learning of policy in the dynamics model. We aim to learn a realistic dynamics model, treating it as a data-driven simulator. Then extensive explorations and evaluations of the agent can occur in $\hat{T}$, thereby reducing the reliance on real-world samples.

\section{Method}
\begin{figure*}[pt!]
    \centering
    \includegraphics[width=\linewidth]{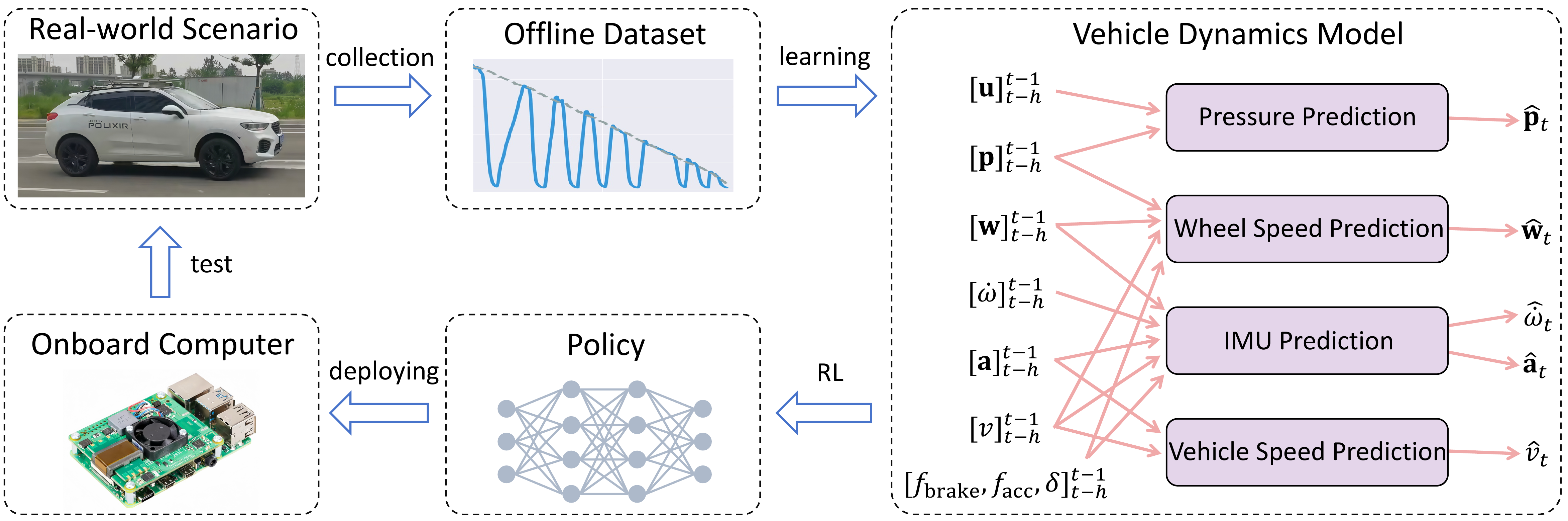}
    \caption{Illustration of ReinVBC: vehicle braking controller by offline model-based reinforcement learning}
    \label{scheme}
\end{figure*}

In this section, we propose ReinVBC, which applies an offline model-based reinforcement learning approach to deal with the vehicle braking control task. The overall scheme of ReinVBC can be divided into five parts: RL environment setup, data collection, vehicle dynamics model learning, policy optimization, and functional test, as illustrated in Figure \ref{scheme}.

\subsection{RL Environment Setup}

\begin{table}[h!]
    \centering
    \caption{Parameters of our experimental vehicle}
    \label{veh_param}
    \begin{tabular}{c|ll}
        \toprule
         \multirow{9}{*}{\textbf{Body}}
         & class & SUV \\
         & length & 4462mm \\
         & width & 1857mm \\
         & height & 1638mm \\
         & wheelbase & 2680mm \\
         & clearance & 170mm \\
         & curb weight & 1630kg \\
         & doors / seats & 5 / 5 \\
         & fuel tank & 55L \\
         \midrule
         \multirow{6}{*}{\textbf{Engine}}
         & class & 2.0T 227 hp L4 \\
         & engine model & E20CB \\
         & displacement & 1499cc \\
         & engine power & 227hp \\
         & maximum torque & 387Nm \\
         & maximum speed & 190km/h \\
         \midrule
         \multirow{3}{*}{\textbf{Drivetrain}}
         & class & DCT \\
         & gears & 7 \\
         & drivetrain type & FWD \\
         \midrule
         \multirow{2}{*}{\textbf{Wheel}}
         & tires & 225 / 60 R18 \\
         & wheel material & aluminum alloy \\
        \bottomrule
    \end{tabular}
\end{table}

\subsubsection{Environment Setting}
We consider several real-world scenarios, including emergency braking on high-adhesion, low-adhesion, and split-friction straight roads. We take a section of concrete road as a high-adhesion surface for our experiments. Certain areas of this concrete road are paved with oil-coated plastic sheets to serve as low-adhesion surfaces. We can simulate a split-friction road condition by having one side of the vehicle's wheels travel on the high-adhesion surface and the other on the low-adhesion surface. 

We use an SUV (Sport Utility Vehicle) named WEY VV5 as the experimental vehicle. Its detailed parameters are listed in Table \ref{veh_param}. The vehicle is equipped with a series of sensors capable of acquiring the values of dynamic variables during driving.

\subsubsection{State Space}
The important vehicle dynamics variables listed in Table \ref{var} are observable to the agent. We define the observation as $o_t\coloneqq (v_t,\mathbf{w}_t,\mathbf{a}_t,\dot{\mathbf{\omega}}_t,\mathbf{p}_t,f_{\mathrm{brake},t},f_{\mathrm{acc},t},\delta_t)$, where $t$ indicates the index of time step. However, these observable variables do not fully describe the vehicle's state. To make the scenario as Markovian as possible, we stack observations from several previous steps to form the state as 
\begin{equation}
s_t\coloneqq([v]_{t-h+1}^t,[\mathbf{w}]_{t-h+1}^t,[\mathbf{a}]_{t-h+1}^t,[\dot{\mathbf{\omega}}]_{t-h+1}^t,[\mathbf{p}]_{t-h+1}^t,[f_{\mathrm{brake}}]_{t-h+1}^t,[f_{\mathrm{acc}}]_{t-h+1}^t,[\delta]_{t-h+1}^t),
\end{equation}
where $h$ is the number of stacked steps.

It is worth mentioning that due to the varying conditions of the road surface during emergency braking of the vehicle, $s_t$ actually lacks information. Considering the difficulty in standardizing the quantitative assessment of various road surfaces and the impracticality of obtaining the road condition in real applications, it is reasonable that $s_t$ does not include road-related information. The trend of changes in dynamic variables contained in the stacking of $h$ steps indirectly reflects the road surface information to some extent, which also tests the model’s ability to extract information.

\begin{table*}[t!]
    \centering
    \caption{Discrete action choices for each wheel.}
    \resizebox{\linewidth}{!}{
    \begin{tabular}{c|ll}
        \toprule
        \textbf{action} & \textbf{description} & \textbf{operation} \\
        \midrule
        $c_0$ & pressure maintenance & set the switching valve to braking mode, close the inlet valve and the outlet valve\\
        $c_1$ & pressure increase & set the switching valve to braking mode, open the inlet valve, close the outlet valve\\
        $c_2$ & pressure decrease & set the switching valve to braking mode, close the inlet valve, open the outlet valve\\
        $c_3$ & no pressure control & set the switching valve to normal mode\\
        \bottomrule
    \end{tabular}
    }
    \label{act_def}
\end{table*}

\subsubsection{Action Space}
We denote the action as $u_t=(u^1_t,u^2_t,u^3_t,u^4_t)$, where $u^i_t$ is the action of the $i$-th wheel. We design four discrete action choices for each wheel. The total number of action choices is 256 at each step. The meaning of each action is specified in Table \ref{act_def}. These actions switch switching valves, inlet valves, and outlet valves of the hydraulic control unit \cite{hcu} to control the wheel cylinder pressure.

Due to the limitations of the sensor frequency, the feedback cycle of $s_t$ is at least 20ms. Therefore, we set the control frequency to 50Hz as well. Considering factors such as sensor delay and drastic state transitions during vehicle braking, our control model lacks sufficient granularity at a low control frequency of 50Hz, posing a significant challenge for RL.

\subsection{Data Collection}
We use a rule-based policy and a random policy to collect data. The rule-based policy, determining pressure control based on the slip ratio, chooses the action for the $i$-th wheel by
\begin{equation}
u^i = c_3\cdot\mathbb{I}(\eta^i<0.03) + c_1\cdot\mathbb{I}(0.03\leq\eta^i<0.1)+ c_0\cdot\mathbb{I}(0.1\leq\eta^i<0.2) + c_2\cdot\mathbb{I}(\eta^i\geq0.2).
\end{equation}
The random policy uniformly samples an action from the action space to execute. Each policy collects six trajectories (five for training, one for validation) on high-adhesion, low-adhesion, and split-friction straight roads, resulting in 36 trajectories in total. The sampling frequency and control frequency are both set to 50Hz. The braking speed during data collection is 40km/h to ensure safety. Besides, the dataset only includes data at lower braking speeds than the test phase, which helps evaluate the method's generalization.

\subsection{Vehicle Dynamics Model Learning}
The vehicle dynamics model is designed to predict dynamic variables, including wheel cylinder pressure, wheel speed, acceleration, attitude rate, and vehicle speed. The right part of Figure \ref{scheme} illustrates the causal graph of the entire dynamics model, indicating that the dynamics model can be decomposed into four modules. Each module only takes the attribution of the dynamic variable it aims to predict as input. To process the input sequence of $h$ steps, the network architecture of each module consists of an RNN \cite{rnn} layer with a GRU \cite{gru} cell and three fully connected layers. The output of each module is a differentiable Gaussian distribution of the predicted dynamic variable.

We denote the whole vehicle dynamics model as $\hat{T}_\xi(d_{t+1}|s_t,u_t)$, where $d_{t+1}$ is used to denote the dynamic variables $(\mathbf{p}_{t+1},\mathbf{w}_{t+1},\dot{\omega}_{t+1},\mathbf{a}_{t+1},v_{t+1})$ and $\xi$ represents the neural parameters. During the roll-out process in the dynamics model, the dynamic $\hat{d}_{t+1}$ is from the model prediction, while the static variables $(f_{\mathrm{brake},t+1}, f_{\mathrm{acc},t+1},\delta_{t+1})$ are from the dataset, collectively used to form the next state $\hat{s}_{t+1}$. $\hat{T}_{\xi}$ is trained to maximize the expected likelihood:
\begin{equation}
\label{model_obj}
J_{\hat{T}}(\xi)=\mathbb{E}_{\mathcal{D}_{\mathrm{env}}}\big[\log\hat{T}_\xi(d_{t+1}|s_t,u_t)+\sum_{k=2}^m\log\hat{T}_\xi(d_{t+k}|\hat{s}_{t+k-1},u_{t+k-1})\big],
\end{equation}
where $m$ is the maximum roll-out length. The first term $\log\hat{T}_\xi(d_{t+1}|s_t,u_t)$ aims to reduce the single-step prediction error, while the purpose of the second term $\sum_{k=2}^m\log\hat{T}_\xi(d_{t+k}|\hat{s}_{t+k-1},u_{t+k-1})$ is to reduce the roll-out error. The hyper-parameters for vehicle dynamics model learning are listed in Appendix \ref{model_hyper}.

\begin{algorithm*}[t!]
    \caption{ReinVBC: MBRL approach to vehicle braking control}
    \label{reinvbc_code}

    \textbf{Input}: Pre-collected real-world dataset $\mathcal{D}_{\mathrm{env}}$, initial vehicle dynamics model $\hat{T}_\xi$, critic $Q_\zeta$ and policy $\pi_\chi$, observation stacking steps $h$, model data buffer $\mathcal{D}_\mathrm{model}$, speed augmentation range $[l_{\mathrm{aug}}, h_{\mathrm{aug}}]$, terminated vehicle speed $v_\epsilon$, interaction epochs $N$, episodes per epoch $E$, maximum time steps per episode $H_\mathrm{max}$
    
    \begin{algorithmic}[1]
        \STATE Train the vehicle dynamics model $\hat{T}_\xi$ on $\mathcal{D}_\mathrm{env}$ by maximizing Equation \eqref{model_obj} to convergence
        \FOR{$N$ epochs}
            \FOR{$E$ episodes}
                \STATE Sample $([v]_{t_0-h+1}^{t_0},[\mathbf{w}]_{t_0-h+1}^{t_0},[\mathbf{a}]_{t_0-h+1}^{t_0},[\dot{\mathbf{\omega}}]_{t_0-h+1}^{t_0},[\mathbf{p}]_{t_0-h+1}^{t_0},[f_{\mathrm{brake}}]_{t_0-h+1}^{t_0},[f_{\mathrm{acc}}]_{t_0-h+1}^{t_0},[\delta]_{t_0-h+1}^{t_0})$, \emph{i.e.}, $h$-step observations prior to step $t_0$, from $\mathcal{D}_\mathrm{env}$ to initial the state
                \STATE Sample a speed augmentation term $v_{\mathrm{aug}}$ uniformly from $[l_{\mathrm{aug}}, h_{\mathrm{aug}}]$

                \FOR{$t=t_0$ to $H_{\mathrm{max}}$ \textbf{if} $v_t>v_\epsilon$}
                \STATE Sample action $a_t$ according to $\pi_\chi(\cdot|s_t)$
                \STATE Perform $a_t$ in the vehicle dynamics model $\hat{T}_\xi$ to predict the state $s_{t+1}$ at next step
                \STATE Compute the reward $r_t$ according to Equation \eqref{r_def}
                \STATE Augment the data by regarding the state as \\
                $s_t^\mathrm{aug}=([v]_{t-h+1}^{t}+v_{\mathrm{aug}},[\mathbf{w}]_{t-h+1}^{t}+v_{\mathrm{aug}},[\mathbf{a}]_{t-h+1}^{t},[\dot{\mathbf{\omega}}]_{t-h+1}^{t},[\mathbf{p}]_{t-h+1}^{t},[f_{\mathrm{brake}}]_{t-h+1}^{t},[f_{\mathrm{acc}}]_{t-h+1}^{t},[\delta]_{t-h+1}^{t})$
                \STATE Add the augmented imaginary sample $(s_t^\mathrm{aug},a_t,r_t,s_{t+1}^\mathrm{aug})$ to $\mathcal{D}_\mathrm{model}$
                \STATE Update current critic $Q_\zeta$ using samples from $\mathcal{D}_{\mathrm{model}}$ by minimizing Equation \eqref{q_obj}
                \STATE Update current policy $\pi_\chi$ using samples from $\mathcal{D}_{\mathrm{model}}$ by maximizing Equation \eqref{pi_obj}
                \ENDFOR
            \ENDFOR
        \ENDFOR
    \end{algorithmic}
\end{algorithm*}

\subsection{Policy Optimization}
The policy optimization process happens in the learned vehicle dynamics model. We expect the vehicle to avoid body deviation and keep the slip ratio within a specific range during braking while achieving a short braking distance. Thus, the reward function is designed as
\begin{equation}
\label{r_def}
r_t=-\beta_1\cdot \hat{v}_{t+1}-\beta_2\cdot \left|\hat{\dot{\psi}}_{t+1}-\dot{\psi}_{\mathrm{exp},t+1}\right|-\beta_3\cdot\sum_{i=1}^4\left[\mathbb{I}(\eta^i_{t+1}<0.1)+\mathbb{I}(\eta^i_{t+1}>0.2)\right],
\end{equation}
where $\beta_1$, $\beta_2$ and $\beta_3$ are the coefficients, $\dot{\psi}_{\mathrm{exp},t+1}$ is the expected yaw rate which can be estimated by $\frac{v\cdot\tan(\delta/N_s)}{L_{\mathrm{veh}}}$ at low speeds and small steering angles. The first term is the vehicle speed penalty, and its integral corresponds exactly to the braking distance objective. The second term represents the difference between the yaw rate response and the expected yaw rate corresponding to the driver's operation, aimed at avoiding body deviation from the expected trajectory. The third term represents the slip ratio constraint, limiting the action range of the policy.

The policy extensively explores in the learned dynamics model and stores the collected model imaginary samples in $\mathcal{D}_{\mathrm{model}}$. Meanwhile, we utilize the off-policy algorithm SAC \cite{sac} to update the policy using the samples from the buffer $\mathcal{D}_{\mathrm{model}}$. Sample exploration and policy optimization iterate alternately. We denote the critic and the actor as $Q_\zeta(s_t,u_t)$ and $\pi_\chi(u_t|s_t)$ respectively, where $\zeta$ and $\chi$ are the neural parameters. The critic is trained to minimize the soft Bellman residual:
\begin{equation}
\label{q_obj}
J_Q(\zeta)=\mathbb{E}_{\mathcal{D}_\mathrm{model}}\left[\frac{1}{2}\left(Q_\zeta(s_t,u_t)-y_{\bar{\zeta},\chi}(r_t,s_{t+1})\right)^2\right],
\end{equation}
where 
\begin{equation}
y_{\bar{\zeta},\chi}(r_t,s_{t+1})=r_t+\gamma\mathbb{E}_{a^\prime\sim\pi_\chi(\cdot|s_{t+1})}\big[Q_{\bar{\zeta}}(s_{t+1},a^\prime)-\alpha\log\pi_\chi(a^\prime|s_{t+1})\big]
\end{equation}
is the soft Bellman target with $\bar{\zeta}$ representing the neural parameters of the target network and $\alpha$ denoting the coefficient of the entropy. The actor is trained to maximize
\begin{equation}
\label{pi_obj}
J_\pi(\chi)=\mathbb{E}_{s_t\sim\mathcal{D}_\mathrm{model}}\left[\mathbb{E}_{a^\prime\sim\pi_\chi(\cdot|s_t)}\left[q_{\zeta,\chi}(s_t,a^\prime)\right]\right],
\end{equation}
where
\begin{equation}
q_{\zeta,\chi}(s_t,a^\prime)=Q_\zeta(s_t,a^\prime)-\alpha\log\pi_\chi(a^\prime|s_t)
\end{equation}
is the Q value with entropy. The hyper-parameters for policy optimization are listed in Appendix \ref{policy_hyper}. 

Under the influence of the entropy term, the policy considers not only the cumulative discounted reward but also the diversity of actions during the learning process. This encourages the policy to explore more regions while balancing multiple optimal paths, making the learned policy more robust. This is also why most model-based algorithms currently choose SAC \cite{sac} to optimize the policy.

It's worth noting that since the dataset only includes low-speed braking data, the roll-out in the dynamics model would also generate low-speed trajectories. To enable the policy to make decisions at high speeds, we conduct data augmentation. Specifically, for an episode generated by the dynamics model, we add a uniformly random number within a certain range to the speed sequence. This specific data augmentation is actually a distinct advantage of data-driven control methods when dealing with out-of-distribution sensor data.

Overall, after collecting the real-world data, the training process of ReinVBC is as shown in Algorithm \ref{reinvbc_code}.

\subsection{Onboard Deployment and Test}
The trained policy is deployed onto the onboard computer, which receives sensor signals and sends control signals via the CAN (Controller Area Network) port. During testing, data is recorded to estimate the braking distance and vehicle deviation for policy evaluation. The newly collected data from the policy is added to the dataset for the next iteration, continuing until the policy's test performance meets the expectations.

\section{Results}
This section presents the results from the following three aspects: (1) the learning results of the vehicle dynamics model and the braking policy; (2) the performance of the learned policy when tested in scenarios consistent with data collection; (3) the performance of the learned policy when tested in scenarios with higher vehicle speeds and greater difficulty.

\begin{figure*}[t!]
    \centering
    \includegraphics[width=\linewidth]{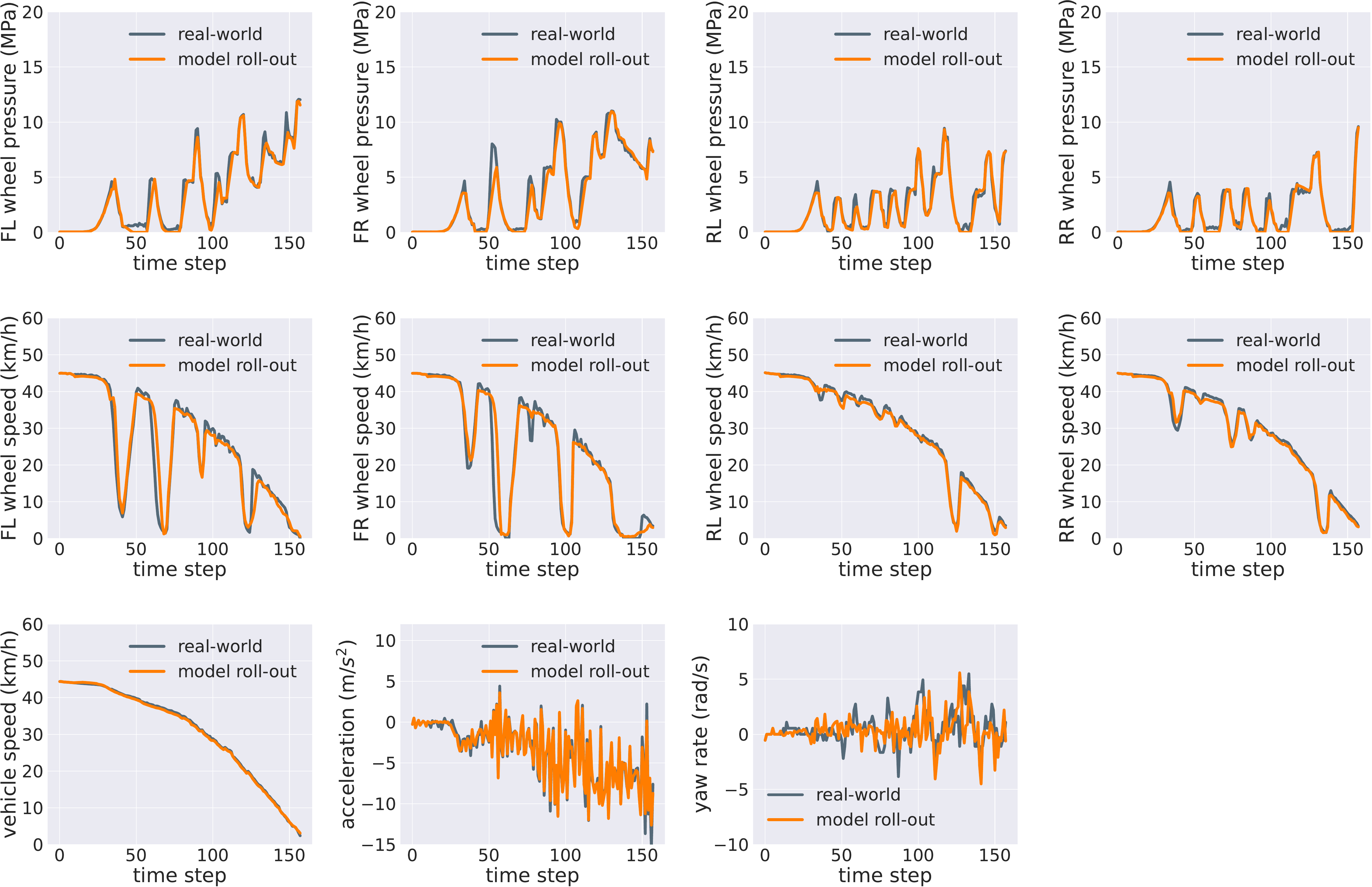}
    \caption{Comparison between the model roll-out (in orange) and the real-world sequence (in blue).}
    \label{model_roll}
\end{figure*}
\begin{figure*}[t!]
    \centering
    \includegraphics[width=\textwidth]{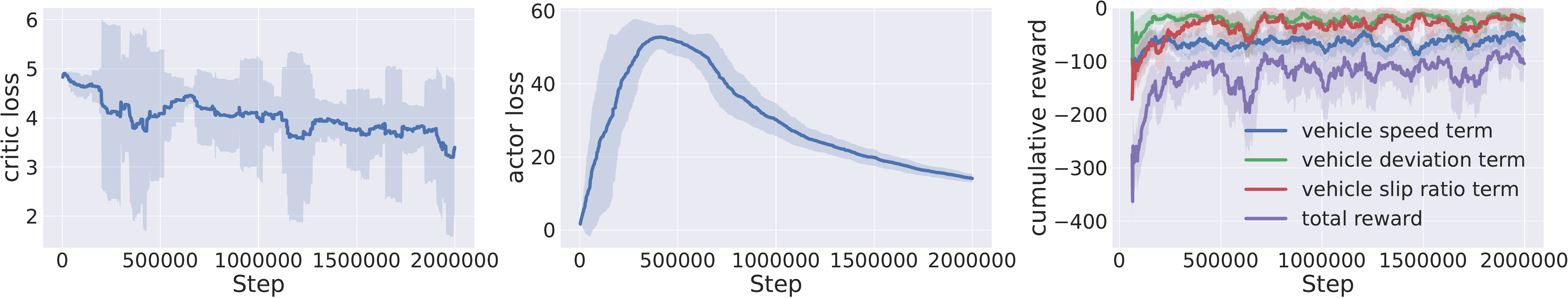}
    \caption{Learning curves of SAC in the learned vehicle dynamics model. The solid lines indicate the mean while the shaded areas indicate the standard error over five different seeds.}
    \label{policy_learn}
\end{figure*}

\subsection{Learning Results}
First, we demonstrate the differences between the learned vehicle dynamics model and the real world. To ensure that the changing trend of the stacked $h$-step observations can adequately reflect the unknown environmental information, we set $h$ to 20, a relatively large number.

We select a trajectory from the validation dataset, start from the initial state, sequentially execute the action sequence in the dynamics model, and record the sequence of dynamic variables generated during the roll-out process. Information related to the steering wheel and pedals controlled by the driver is directly taken from the dataset. The generated sequence from the dynamics model is then compared with the real-world sequence. Figure \ref{model_roll} illustrates the comparison results for the wheel cylinder pressure, wheel speed, vehicle speed, acceleration, and yaw rate of the trajectory from the validation dataset. In the figure, FL (front left), FR (front right), RL (rear left), and RR (rear right) indicate the position of the wheel. More additional results can be found in Appendix \ref{additional_roll}. 

It can be observed that the learned vehicle dynamics model has an outstanding ability to predict future states. Firstly, across all dynamic variables, the two curves closely align and do not exhibit significant gaps as the roll-out length increases, suggesting that the learned model has negligible roll-out error. Secondly, from the roll-out curves of different scenarios included in Figure \ref{model_roll} and Appendix \ref{additional_roll}, it can be seen that the model consistently performs well in prediction across various scenarios, demonstrating strong coverage capability. Furthermore, the model tends to ignore some minor changes, resulting in smoother prediction results compared to the actual data. Learning a policy for braking control is promising in such a dynamics model.

Next, we present the learning curves of policy optimization in the learned vehicle dynamics model. Figure \ref{policy_learn} shows the critic loss curve, the actor loss curve, and the reward curves during the learning process. It can be observed that the loss curves tend to decrease, and each term in the reward function tends to increase. The learning process proceeds successfully without abnormal phenomena like Q-value explosion, even though we do not apply a model-uncertainty penalty \cite{mopo,morel,mobile,any_step} to the samples collected from the dynamics model. This is because the slip ratio term included in the reward function restricts the policy to explore only within the regions where the slip ratio constraints are met. These exploration regions align with those of the rule-based policy used during data collection. In other words, the policy's exploration area largely corresponds to the certain regions of the dynamics model. Additionally, the learning quality of the dynamics model itself is reliable enough to support some out-of-distribution explorations by the policy.

\subsection{In-distribution Test in Real-world}
\begin{figure*}[pt!]
    \centering
    \includegraphics[width=\linewidth]{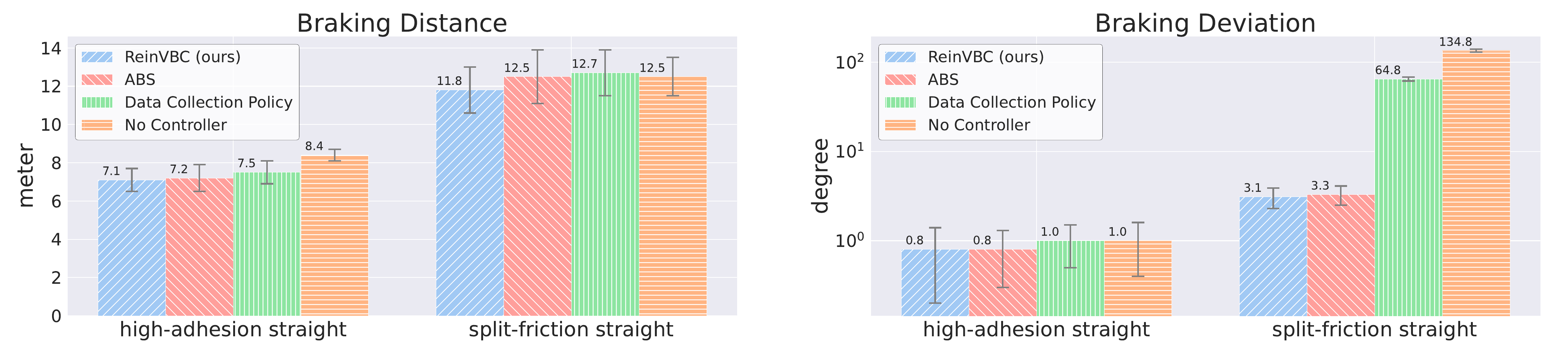}
    \caption{In-distribution test results in real-world (with 40km/h as braking speed), in terms of braking distance (in meter) and braking deviation (in degree). The height of the bars represents the mean, and the error bars indicate the standard deviation, over five experiments. Due to the significant differences in values for braking deviation across different methods, the y-axis of the corresponding chart uses a logarithmic scale.}
    \label{in_test}
\end{figure*}

In the scenarios identical to data collection, keeping the initial vehicle speed at 40km/h for braking, we compare the braking distance and braking deviation of the controller learned by offline MBRL, \emph{i.e.}, ReinVBC, the original-equipment Bosch ABS controller, the rule-based policy used for data collection, and direct braking without any pressure control. Figure \ref{in_test} shows the in-distribution test results. The height of the bars represents the mean, and the error bars indicate the standard deviation, over five experiments. Due to the significant differences in values for braking deviation across different methods, the y-axis of the corresponding chart uses a logarithmic scale.

First of all, ReinVBC demonstrates noticeable improvement over direct braking without any controller in both braking distance and braking deviation. It is worth noting that when braking without a controller on a split-friction road surface, wheel lock-up causes uneven force distribution on the vehicle, resulting in significant deviation. ReinVBC, on the other hand, has the ability to avoid vehicle deviation. This indicates that the RL-learned controller can effectively control the wheel cylinder pressure during intense braking, ensuring the car’s safety and steer-ability.

Next, it is worth focusing on the comparison between ReinVBC and the simple rule-based policy used for data collection. The data collection policy shows almost no improvement compared to direct braking without any control, suggesting its limited control to the wheel cylinder pressure. However, data collected with such a low-performance policy can still be used to learn the effective controller ReinVBC. The fundamental reason is that although the collected sequences are far from the optimal trajectory, the dynamic characteristics reflected in the data can support the learning of the environmental model. Then, using an RL algorithm, it is easy to find better control policy within the environmental model. Although the improvement of the RL-learned policy compared to the behavior policy used for data collection cannot yet be precisely quantified based on specific theory, the performance enhancement is sufficient to support the application of offline MBRL.

What's more, with only 36 trajectories of braking data, ReinVBC can achieve production-level ABS control performance in terms of braking distance and braking deviation. Although the out-performance is slight, ReinVBC requires less than an hour of data collection and a few days of training and hyper-parameter tuning, which is significantly more efficient than the extensive time required for manual calibration to meet production-level ABS standards. This result already suggests the feasibility and potential of offline MBRL in real-world applications. Given additional data collection in more complex scenarios, offline MBRL is likely to demonstrate even more pronounced advantages.

\subsection{Out-of-distribution Test}
The out-of-distribution test is used to assess the generalization ability of the learned policy in challenging scenarios. To ensure the safety, we divide the test into two stages. At the first stage, the test is conducted in a hardware-in-loop simulation, posing no risk. At the second stage, the learned policy is deployed in the real-world, where extremely serious accidents will occur once the policy generates anomalous control signals. Therefore, only when the policy’s decisions in the simulation are sufficiently reasonable and can achieve satisfactory performance will it proceed to the second stage.

\subsubsection{hardware-in-loop Out-of-distribution Test}
At this stage, the control signals generated by the policy are applied to a physical hydraulic control unit to produce wheel cylinder pressure signals, which are then transmitted via a CAN port to the CarSim \footnote{www.carsim.com} software on the computer. CarSim simulates the vehicle’s motion, and upon receiving the pressure signals, the vehicle’s dynamic variables transition to new values, which are then fed back to the policy as observations. We set up seven challenging braking scenarios in CarSim to test the performance of the learned policy and compare it with the ABS controller and direct braking without any pressure control. Table \ref{hil_out} shows the results.

\begin{table*}[t!]
    \centering
    \caption{hardware-in-loop Out-of-distribution Test Results}
    \resizebox{\linewidth}{!}{
    \begin{tabular}{cc|ccc|ccc}
    \toprule
    \multirow{2}{*}{\textbf{Scenario}}
    & \multirow{2}{*}{\textbf{Braking Speed}}
    & \multicolumn{3}{c}{\textbf{Braking Distance (m)}} & \multicolumn{3}{c}{\textbf{Braking Deviation (degree)}}  \\
    \cmidrule{3-8}
    & \textbf{(km/h)}& ReinVBC & ABS & No Controller & ReinVBC & ABS & No Controller \\
    \midrule
    high-adhesion straight & $100$ & $34.9\pm0.5$ & $\mathbf{27.6\pm0.6}$ & $63.3\pm0.3$& $<0.1$ & $<0.1$ & $<0.1$ \\
    low-adhesion straight & $55$ & $65.0\pm0.7$ & $70.7\pm0.5$ & $\mathbf{56.8\pm0.4}$ & $\mathbf{0.3\pm0.1}$ & $\mathbf{0.3\pm0.1}$ & $439.3\pm7.2$\\
    high-to-low straight & $45$ & $\mathbf{30.5\pm0.6}$ & $32.3\pm0.5$& $46.8\pm0.5$ & $\mathbf{0.2\pm0.0}$ &$\mathbf{0.2\pm0.0}$ & $68.4\pm5.1$ \\
    low-to-high straight & $45$ & $\mathbf{25.7\pm0.4}$& $\mathbf{25.6\pm0.5}$ & $41.6\pm0.4$ & $\mathbf{0.2\pm0.0}$ & $\mathbf{0.2\pm0.0}$ & $66.3\pm7.2$ \\
    split-friction straight & $60$ & $\mathbf{53.9\pm3.2}$ & $\mathbf{54.7\pm2.3}$ & $57.5\pm1.4$ & $0.6\pm0.1$ & $\mathbf{0.5\pm0.1}$ & $473.7\pm9.1$\\
    high-adhesion curve & $30$ & $\mathbf{10.2\pm0.5}$ & $26.4\pm0.5$ & $\mathbf{10.0\pm0.7}$ & $\mathbf{0.3\pm0.0}$ & $\mathbf{0.3\pm0.0}$ & $10.9\pm3.7$\\
    split-friction curve & $30$ & $15.8\pm1.6$ & $28.5\pm1.7$ & $\mathbf{11.8\pm1.4}$ & $\mathbf{0.6\pm0.1}$& $0.7\pm0.1$ & $61.9\pm5.4$ \\
    \bottomrule
    \end{tabular}
    }
    \label{hil_out}
\end{table*}

In scenarios other than the high-adhesion straight, direct braking without any pressure control causes the vehicle to deviate significantly from the desired trajectory. The braking deviation caused by wheel lock-up poses a significant threat to driver safety. Although the braking distance may have an advantage in some scenarios without applying control, this shorter braking distance is meaningless when it comes at the cost of steer-ability and safety. In contrast, both ReinVBC and ABS result in negligible braking deviation across all scenarios. It is reasonable to believe that ReinVBC has the capability to ensure risk-free vehicle motion during braking.

\begin{table*}[t!]
    \centering
    \caption{Real-world Out-of-distribution Test Results}
    \begin{tabular}{c|cccc}
    \toprule
    \multirow{2}{*}{\textbf{Scenario}}
     & \textbf{Braking Speed} & \textbf{Braking Distance} & \textbf{Braking Deviation} & \textbf{Wheel Lock-up}  \\
     & \textbf{(km/h)} & \textbf{(m)} & \textbf{(deg)} & \textbf{(Yes/No)}  \\
    \midrule
    high-adhesion straight & $100$ & $47.7\pm1.9$ & $1.7\pm0.8$ & No\\
    low-adhesion straight & $55$ & $57.9\pm2.0$ & $0.9\pm0.2$ & No\\
    high-to-low straight & $45$ & $26.0\pm2.9$ & $0.7\pm0.2$ & No \\
    low-to-high straight & $45$ & $21.9\pm0.8$ & $1.2\pm0.2$ & No\\
    split-friction straight & $60$ & $33.8\pm1.0$ & $8.9\pm0.4$ & No\\
    high-adhesion curve & $30$ & $5.8\pm0.5$ & $0.9\pm0.2$ & No\\
    split-friction curve & $30$ & $9.7\pm0.9$ & $2.7\pm0.3$ & No\\
    \bottomrule
    \end{tabular}
    \label{real_out}
\end{table*}

Moreover, in terms of braking distance, ReinVBC is competitive with ABS in scenarios other than the high-adhesion straight path. Many of these scenarios involve sudden changes in wheel speed, and the observation stacking design of ReinVBC allows it to capture these abrupt changes and make appropriate decisions. However, this capability also makes the policy more cautious, leading to slightly longer braking distances in low-risk scenarios like the high-adhesion straight due to insufficient braking pressure. Nonetheless, this minor flaw is insignificant compared to the stable control it provides in hazardous scenarios. Based on ReinVBC’s excellent performance in hardware-in-loop simulations, it is entirely feasible to further test the learned policy’s performance in the real world.

\subsubsection{Real-world Out-of-distribution Test}
We test ReinVBC in a series of complex scenarios at a professional automotive testing ground. The braking mode is emergency braking with a master cylinder pressure above 16MPa. These scenarios are outside the coverage of the dataset, testing the policy's generalization ability in the real world. Table \ref{real_out} shows the real-world out-of-distribution test results. The curves of the dynamic variables during braking are shown in Appendix \ref{test_seq}. It can be observed that in all seven braking test scenarios, the vehicle with our controller exhibits small braking deviation, indicating the vehicle does not skid and safety is ensured. Additionally, there is no significant wheel lock-up, ensuring steer-ability. These phenomena demonstrate that the policy obtained through offline MBRL is capable of handling emergency braking in common braking scenarios.

The reason for not comparing with the original-equipment ABS controller is that the ABS controller on the experimental vehicle has already been replaced with our controller during the experiments at the professional automotive testing ground. Of course, given the limited training data, it is likely that our controller is not able to compete with the production-grade ABS in out-of-distribution scenarios. The focus of these experiments is to demonstrate the potential of offline MBRL in real-world braking control applications.

\subsection{Reality Gap Analysis}
\begin{figure*}[t!]
    \centering
    \includegraphics[width=\textwidth]{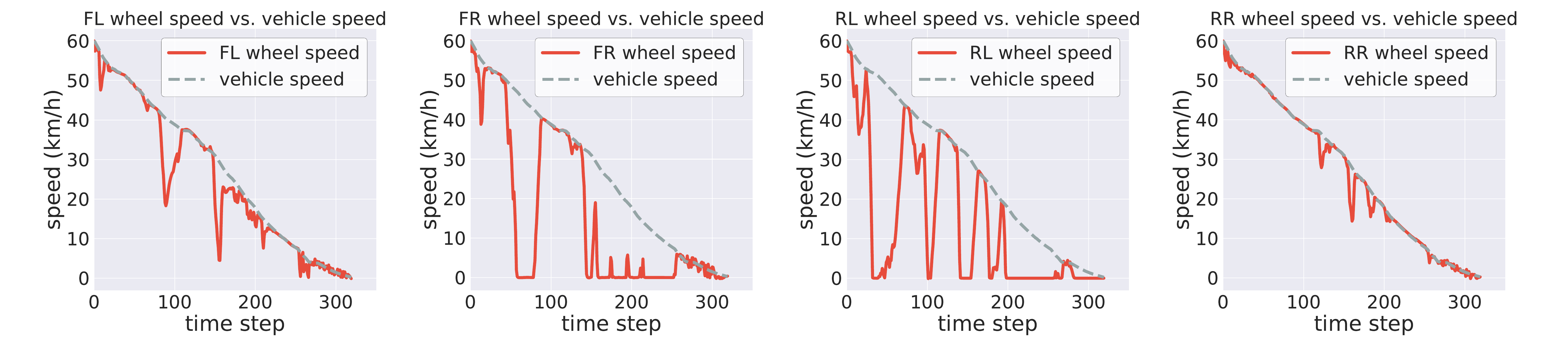}
    \caption{Speed curves during braking on the split-friction straight in the hardware-in-loop simulation. We separately compare the vehicle speed sequence with each wheel’s speed sequence. The solid lines indicate the wheel speed curves of each wheel, while the dashed line indicates the vehicle speed for reference.}
    \label{sim_curve}
\end{figure*}
\begin{figure*}[t!]
    \centering
    \includegraphics[width=\textwidth]{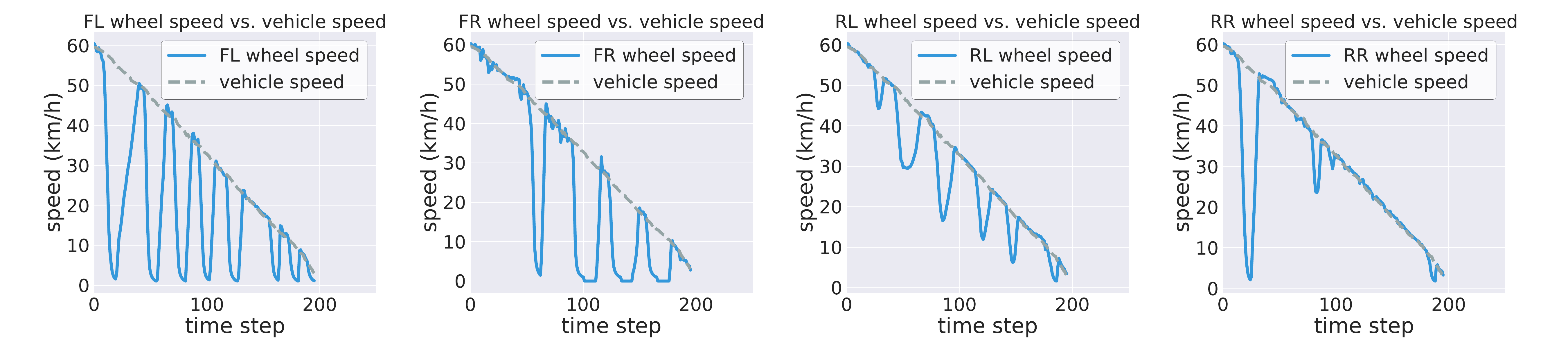}
    \caption{Speed curves during braking on the split-friction straight in the real world. We separately compare the vehicle speed sequence with each wheel’s speed sequence. The solid lines indicate the wheel speed curves of each wheel, while the dashed line indicates the vehicle speed for reference.}
    \label{real_curve}
\end{figure*}

As previously mentioned, to ensure safety, we first validate the performance of the learned policy in a hardware-in-loop simulation before testing it in the real world. Although integrating the physical hydraulic control unit into the simulation loop helps, a reality gap still exists. The span of this reality gap directly determines the feasibility of this validation approach. Therefore, we will proceed with an appropriate analysis of the reality gap.

We select a typical scenario presented in the previous test results, the split-adhesion straight. During the testing of the learned policy in both hardware-in-loop simulations and real-world environments, we record the time-series data of the braking process. We separately present the wheel speed and vehicle speed curves on different testing platforms, as shown in Figure \ref{sim_curve} and Figure \ref{real_curve}.

It can be observed that the braking duration in the hardware-in-loop simulation is longer, and wheel lock-up is more pronounced. The reason is that the low-adhesion surface used at the automotive testing ground is made of basalt, while the CarSim simulation uses an ideal, smoother ice surface. In fact, the ideal low-adhesion surface in the simulation is more rigorous for testing braking controllers. The challenges posed by the braking scenarios in the simulation are no less demanding than those presented in the real-world scenarios set up at the professional automotive testing ground.

More importantly, the wheel speed curves reflect the control logic of the braking controller. Whether in the simulation or the real world, the wheel speed tends to rebound as it approaches zero. This is because when a wheel is about to lock up, the controller reduces the corresponding wheel cylinder pressure to prevent dangerous skidding and vehicle deviation. The relationship between pressure and wheel speed can be seen in the curves in Appendix \ref{test_seq}. Of course, the rebound tendency of the wheel speed is weaker in the simulation, as the smoother low-adhesion surface provides greater resistance to wheel speed rebound.

\begin{table*}[t!]
    \centering
    \caption{Performance Comparison of Testing in Simulation and Real World}
    \resizebox{\linewidth}{!}{
    \begin{tabular}{cc|ccc|ccc}
    \toprule
    \multirow{2}{*}{\textbf{Scenario}}
    & \multirow{2}{*}{\textbf{Braking Speed}}
    & \multicolumn{3}{c}{\textbf{Braking Distance (m)}} & \multicolumn{3}{c}{\textbf{Braking Deviation (degree)}}  \\
    \cmidrule{3-8}
    & \textbf{(km/h)}& Simulation & Real World & Gap & Simulation & Real World & Gap \\
    \midrule
    high-adhesion straight & $100$ & $34.9$ & $47.7$ & $+12.8$ & $<0.1$ & $1.7$ & $+1.6$ \\
    low-adhesion straight & $55$ & $65.0$ & $57.9$ & $-7.1$ & $0.3$ &  $0.9$ & $+0.6$\\
    high-to-low straight & $45$ & $30.5$ & $26.0$& $-4.5$& $0.2$ & $0.7$ & $+0.5$ \\
    low-to-high straight & $45$ & $25.7$& $21.9$& $-3.8$ & $0.2$ & $1.2$ & $+1.0$ \\
    split-friction straight & $60$ & $53.9$ & $33.8$ & $-20.1$ & $0.6$ & $8.9$ & $+8.3$ \\
    high-adhesion curve & $30$ & $10.2$ & $5.8$ & $-4.4$ & $0.3$ & $0.9$ & $+0.6$\\
    split-friction curve & $30$ & $15.8$ & $9.7$ & $-6.1$ & $0.6$& $2.7$ & $+2.1$ \\
    \bottomrule
    \end{tabular}
    }
    \label{sim_real_diff}
\end{table*}

On the other hand, we list the performance comparison of the learned policy tested in simulation and the real world in Table \ref{sim_real_diff}. In terms of braking distance, the gap between simulation and the real world is relatively small in scenarios other than high-adhesion straight and split-adhesion straight. Sources of deviation include vehicle parameters, road conditions, weather conditions, and air resistance. In practice, deviations in braking distance are not a significant issue. Even if braking distances in the real world are longer than in CarSim, it does not pose a direct safety threat. More important is the consistency of braking deviation, as this metric directly reveals the steer-ability of the braking process. It can be observed that the differences in braking deviation between simulation and the real world are not significant. Based on our experimental results, if the policy does not cause noticeable vehicle deviation during braking in simulation, it will not do so in the real world either.

In summary, when conducting risky out-of-distribution testing of a learned policy, if the control logic of the learned policy in simulation is reasonable and key metrics pass, it can proceed to further testing in the real world. Our experimental process can serve as a reference for applying reinforcement learning in other control domains.

\section{Limitations}
In general, our work's limitations are summarized as follows.
\begin{itemize}
    \item This work lacks an all-round comparison between ReinVBC and the production-grade ABS in the real-world testing ground since we only have one experimental vehicle and the original-equipment ABS has been replaced with our controller during test.
    \item The performance of our controller at extremely high vehicle speeds above 100km/h is unclear since it is hard to ensure safety.
    \item This work deploys the learned policy on the onboard computer. The braking policy should be deployed on the micro chip of the vehicle during production. How to utilize the learned neural parameters with limited computing resources should be considered.
\end{itemize}

\section{Related Work}
This work is related to vehicle braking control and offline MBRL.

\subsection{Vehicle Braking Control}
ABS \cite{abs1,abs2,abs3,abs4} is widely used to control the wheel cylinder pressure to maintain steer-ability and safety of the vehicle during heavy braking. Several works make efforts to apply RL to vehicle braking control. Mantripragada and Kumar \cite{rl_brake2} improve the ABS by proposing a model-free RL control algorithm which can adapt to changing tire characteristics and thereby effectively utilizing the available grip at tire-road interface. Radac and Precup \cite{rl_brake3, rl_brake4} apply a model-free Q-Learning for a fast and highly nonlinear ABS. Abreu et al. \cite{rl_brake5} utilize a double deep Q-network to deal with the rough terrain. Sardarmehni and Heydari \cite{rl_brake6} select a value iteration algorithm to offline learn the infinite horizon solution for optimal control of ABS in ground vehicles. But these works still remain in physical simulators or hardware-in-loop simulations. In contrast, our work focuses on the real-world vehicle braking scenario.

\subsection{Offline MBRL}
The core issue of offline MBRL lies in how to effectively learn and leverage the model. To let the model-generated sample be more reliable, several works try to enhance the paradigm of dynamics model learning. For instance, ADMPO \cite{any_step,adm2} proposes an any-step dynamics model, and MOREC \cite{morec} designs a reward-consistent dynamics model using an adversarial discriminator. 

The mainstream works \cite{mopo,morel,mobile,combo,rambo,cbop} focus on the ensemble dynamics model \cite{pets,mbpo} and leverage the learned model conservatively. Concretely, MOPO \cite{mopo} and MOReL \cite{morel} add the uncertainty of the model prediction as a penalization term to the original reward function with the purpose of achieving a pessimistic value estimation. MOBILE \cite{mobile} improves the uncertainty quantification by introducing Model-Bellman inconsistency into the offline model-based framework. COMBO \cite{combo} applies CQL \cite{cql} to force the estimated state-action value to be small on model-generated out-of-distribution samples. RAMBO \cite{rambo} achieves conservatism by adversarial model learning for value minimization while keeping fitting the transition function. CBOP \cite{cbop} introduces adaptive weighting of short-horizon roll-out into MVE \cite{mve} technique and adopts the variance of values under an ensemble of dynamics models to estimate the Q value conservatively.

Our work also applies the framework of offline MBRL but views the learned vehicle dynamics model as a data-driven simulator and makes full-length roll-outs.

\section{Conclusion}
In this paper, we focus on reducing the consumption of labor and time while maintaining the controller performance during the production of vehicular braking system. We apply the framework of offline MBRL, which is a promising solution for addressing real-world control tasks. After modeling the braking task as a Markov decision process, we learn a vehicle dynamics model using the collected offline dataset. Then, we regard the learned dynamics model as a data-driven simulator and optimize the braking policy in it. By deploying the learned policy on the experimental vehicle to control the wheel cylinder pressure during braking, several results show that our method can achieve a small braking deviation and avoid the wheel lock-up in emergency-braking scenarios, ensuring safety and steer-ability of the vehicle, even in scenarios outside the coverage of the dataset. Although the braking controller by offline MBRL is not able to outperform the production-grade ABS absolutely at present, this paper has demonstrated the potential of offline MBRL in real-world vehicle braking. We expect to replace the manual calibration of the traditional controller with a reliable data-driven learning paradigm, reducing the cost of production. 

\bibliographystyle{abbrv}
\bibliography{ref}

\newpage
\appendix
\section{Experimental Details}
\subsection{Hyper-parameters for Vehicle Dynamics Model Learning}
\label{model_hyper}
\begin{table}[ht!]
    \centering
    \fontsize{9}{9}\selectfont
    \caption{Hyper-parameters used to train our vehicle dynamics model}
    \begin{tabular}{lll}
        \toprule
        \textbf{Hyper-parameter} & \textbf{Value} & \textbf{Description} \\
        \midrule
         network & GRU(128)+FC(128,128) & a GRU layer followed by fully connected layers\\
         $h$ & 20 & the number of steps for observation stacking \\
         $p_{\mathrm{dropout}}$ & 0.1 & dropout rate \\
         $lr_{\mathrm{model}}$ & $1\times 10^{-4}$ & learning rate \\
         $m$ & 500 & maximum roll-out length \\
         optimizer & Adam & optimizer of the dynamics model \\
         $N$ & 1000 & number of training epochs \\
         batch size & 128 & batch size for each update \\
        \bottomrule
    \end{tabular}
\end{table}

\subsection{Hyper-parameters for Policy Optimization}
\label{policy_hyper}
\begin{table}[ht!]
    \centering
     \fontsize{9}{9}\selectfont
    \caption{Hyper-parameters used to optimize the policy}
    \begin{tabular}{lll}
        \toprule
        \textbf{Hyper-parameter} & \textbf{Value} & \textbf{Description} \\
        \midrule
        $N_Q$ & 2 & the number of Q networks \\ 
        actor network & FC(256,256) & fully connected (FC) layers\\
        critic network & FC(256,256) & fully connected (FC) layers\\
        $\tau$ & $5\times 10^{-3}$ & target network smoothing coefficient\\
        $\gamma$ & 0.99 & discount factor\\
        $lr_{\mathrm{policy}}$ & $3\times 10^{-4}$ & learning rate of the actor and the critic\\
        optimizer & Adam & optimizer of the actor and the critic \\
        batch size & 256 & batch size for each update\\
        $(\beta_1,\beta_2,\beta_3)$ & (0.025, 0.5, 0.2) & coefficients of each term in the reward function \\
        \bottomrule
    \end{tabular}
\end{table}

\newpage
\section{Additional Results}
\subsection{Roll-out Demonstration of the Vehicle Dynamics Model}
\label{additional_roll}

\begin{figure*}[h!]
    \centering
    \includegraphics[width=0.85\linewidth]{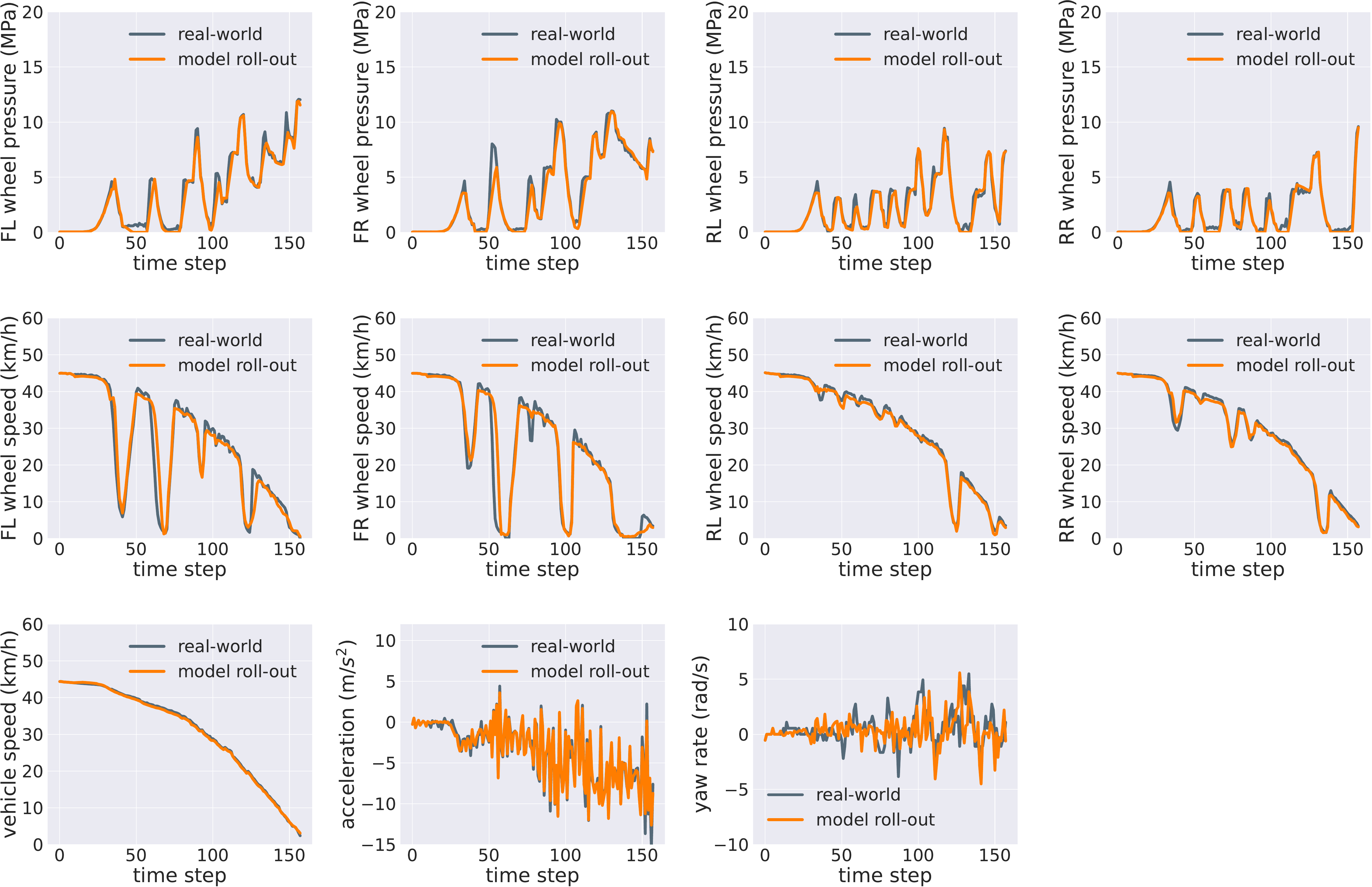}
    \caption{Comparison between the model roll-out and the real-world sequence. This sequence is sampled on a split-friction straight road.}
\end{figure*}

\begin{figure*}[h!]
    \centering
    \includegraphics[width=0.85\linewidth]{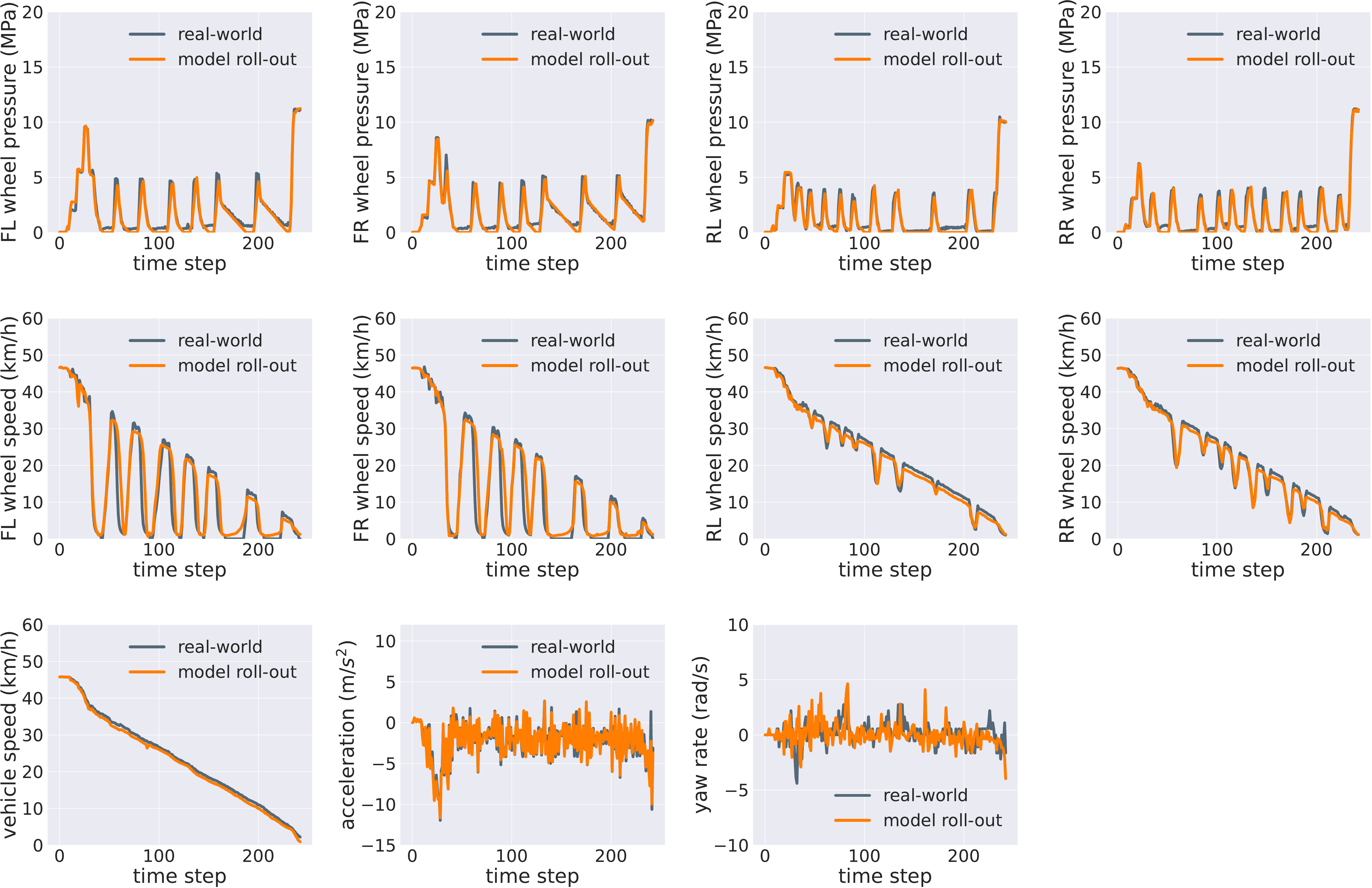}
    \caption{Comparison between the model roll-out and the real-world sequence. This sequence is sampled on a split-friction straight road.}
\end{figure*}

\begin{figure*}[h!]
    \centering
    \includegraphics[width=0.88\linewidth]{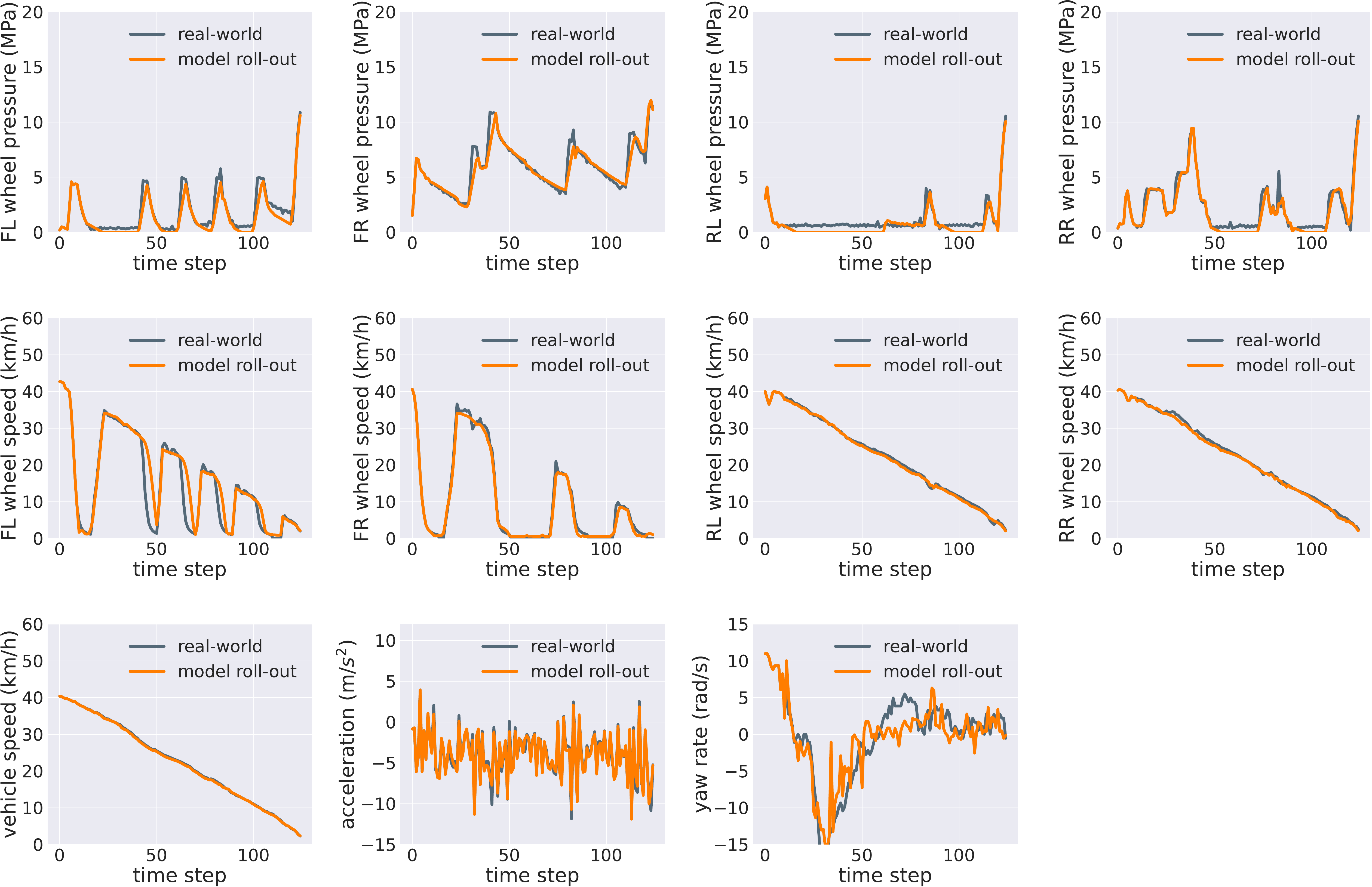}
    \caption{Comparison between the model roll-out and the real-world sequence. This sequence is sampled on a split-friction straight road.}
\end{figure*}

\begin{figure*}[h!]
    \centering
    \includegraphics[width=0.88\linewidth]{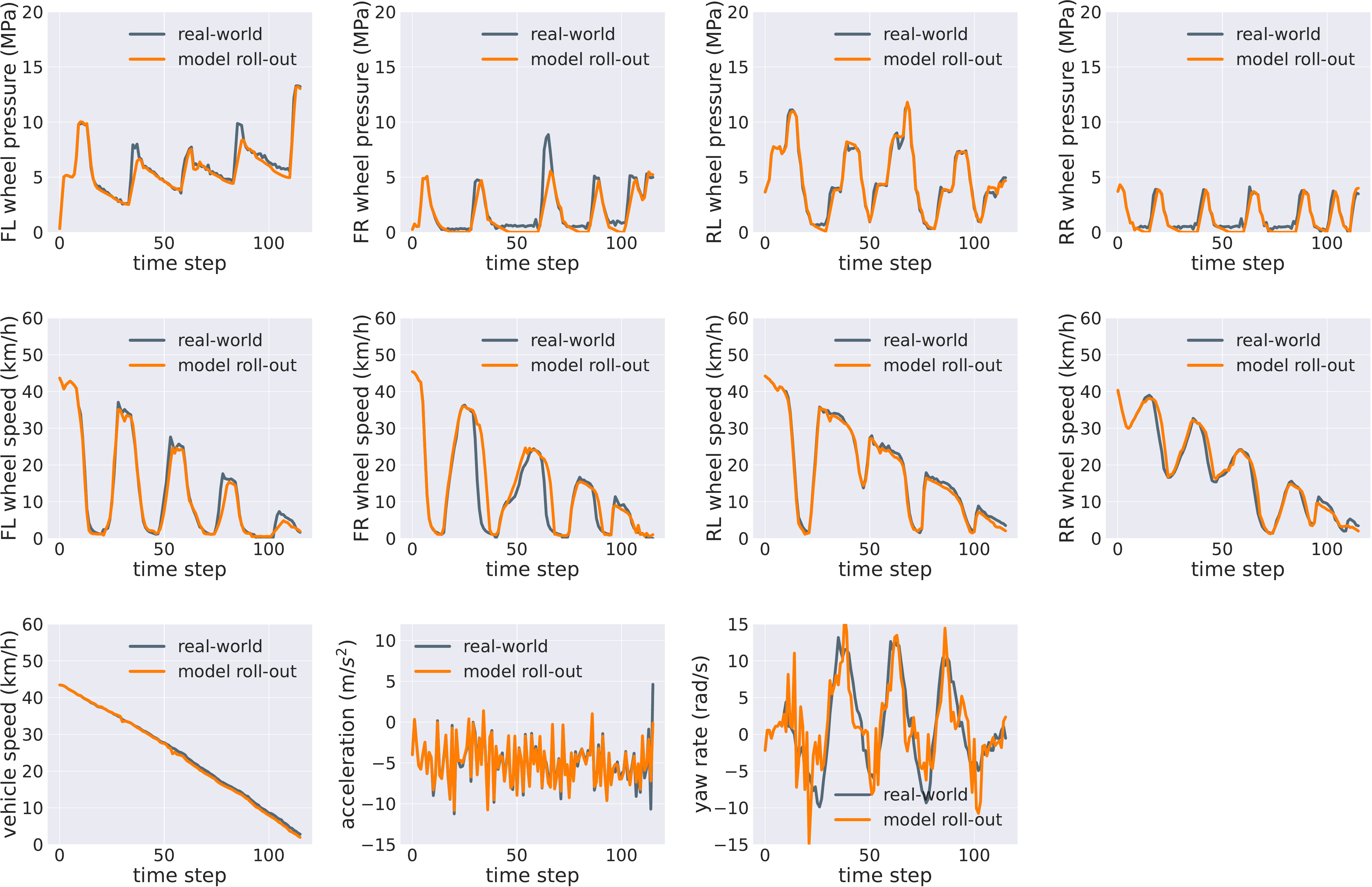}
    \caption{Comparison between the model roll-out and the real-world sequence. This sequence is sampled on a split-friction straight road.}
\end{figure*}

\begin{figure*}[h!]
    \centering
    \includegraphics[width=0.88\linewidth]{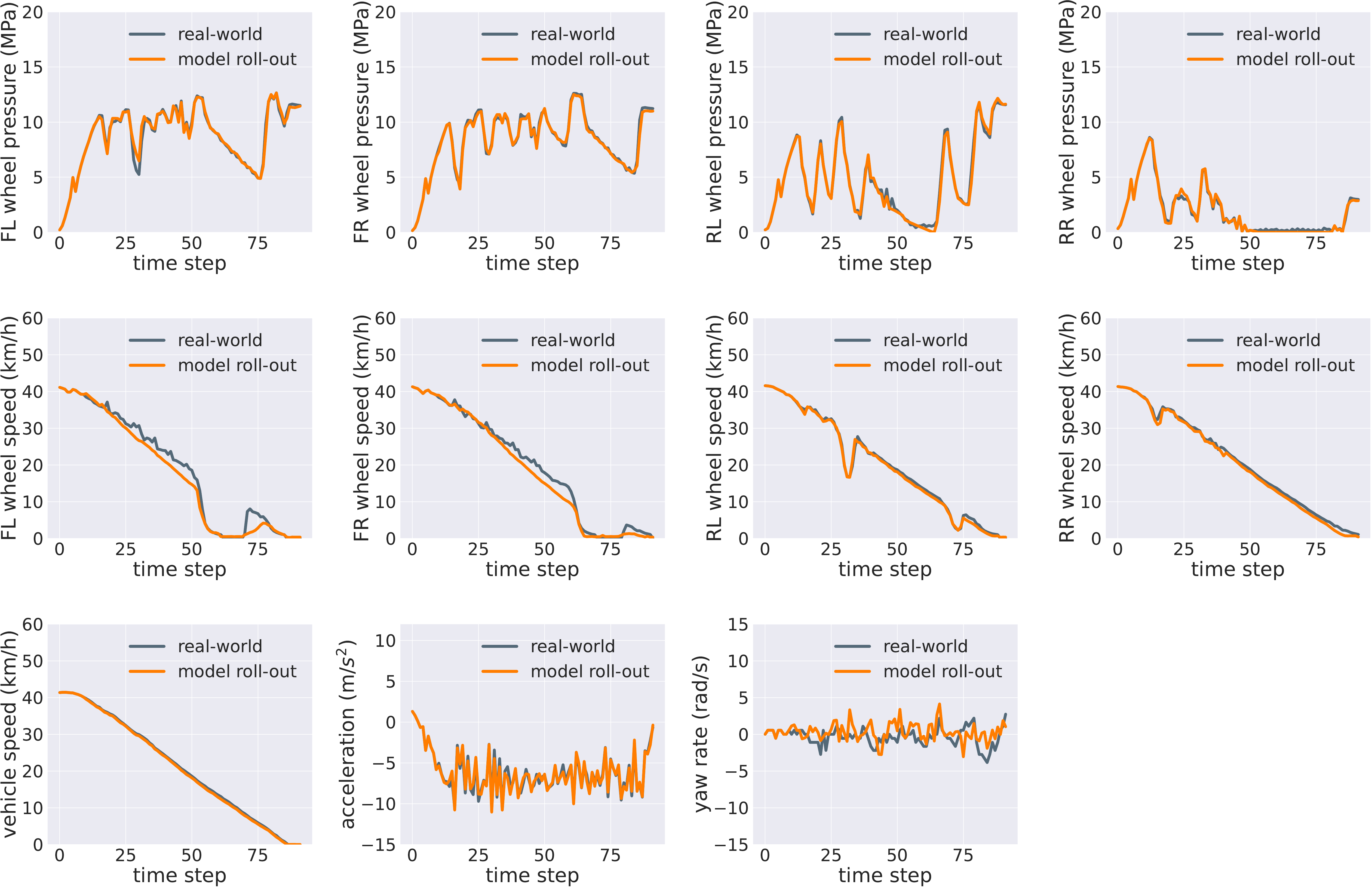}
    \caption{Comparison between the model roll-out and the real-world sequence. This sequence is sampled on a high-adhesion straight road.}
\end{figure*}

\begin{figure*}[h!]
    \centering
    \includegraphics[width=0.88\linewidth]{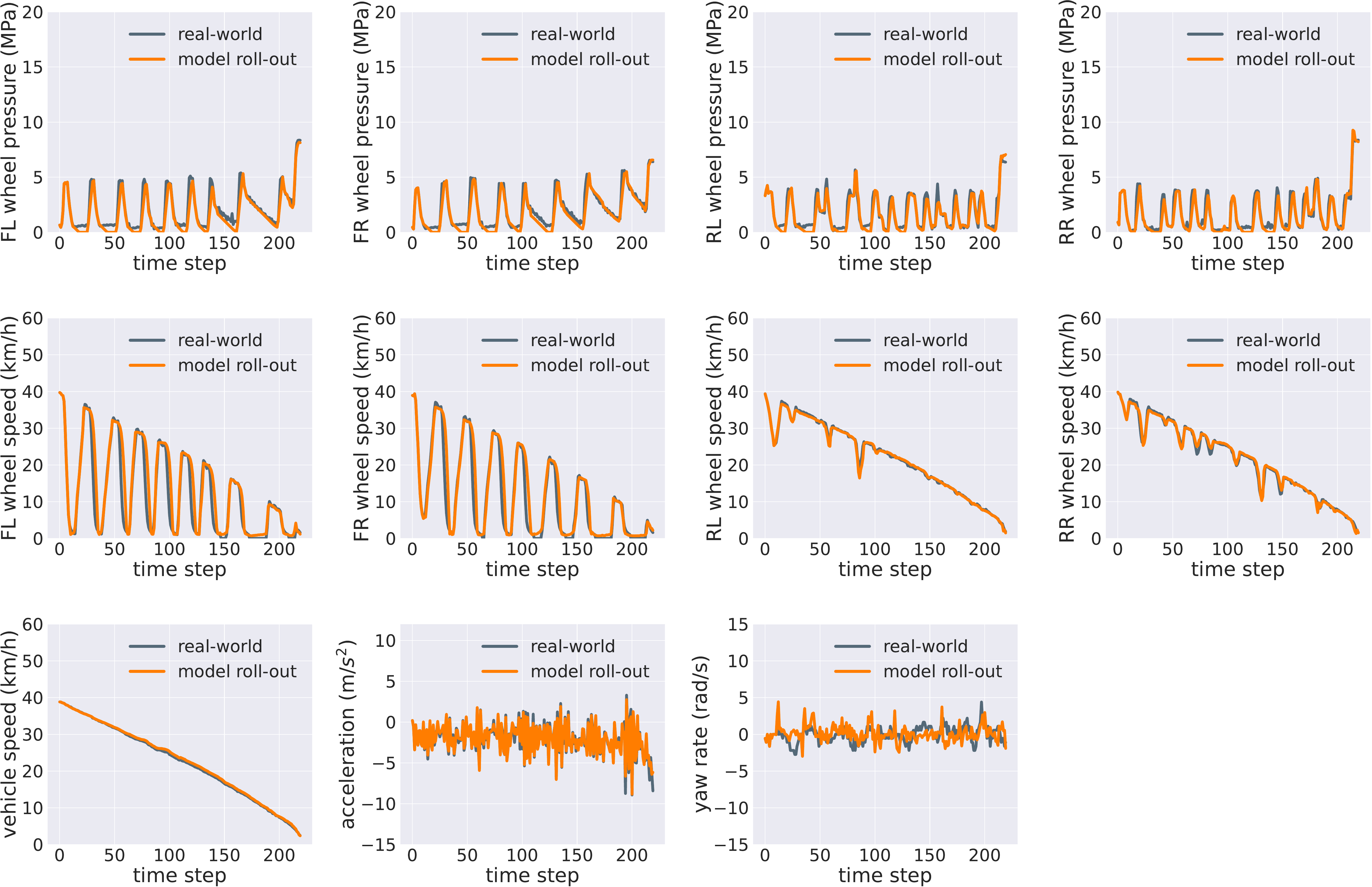}
    \caption{Comparison between the model roll-out and the real-world sequence. This sequence is sampled on a low-adhesion straight road.}
\end{figure*}

\clearpage
\subsection{Test Sequence of the Vehicle Braking Controller}
\label{test_seq}

\begin{figure*}[h!]
    \centering
    \includegraphics[width=0.88\linewidth]{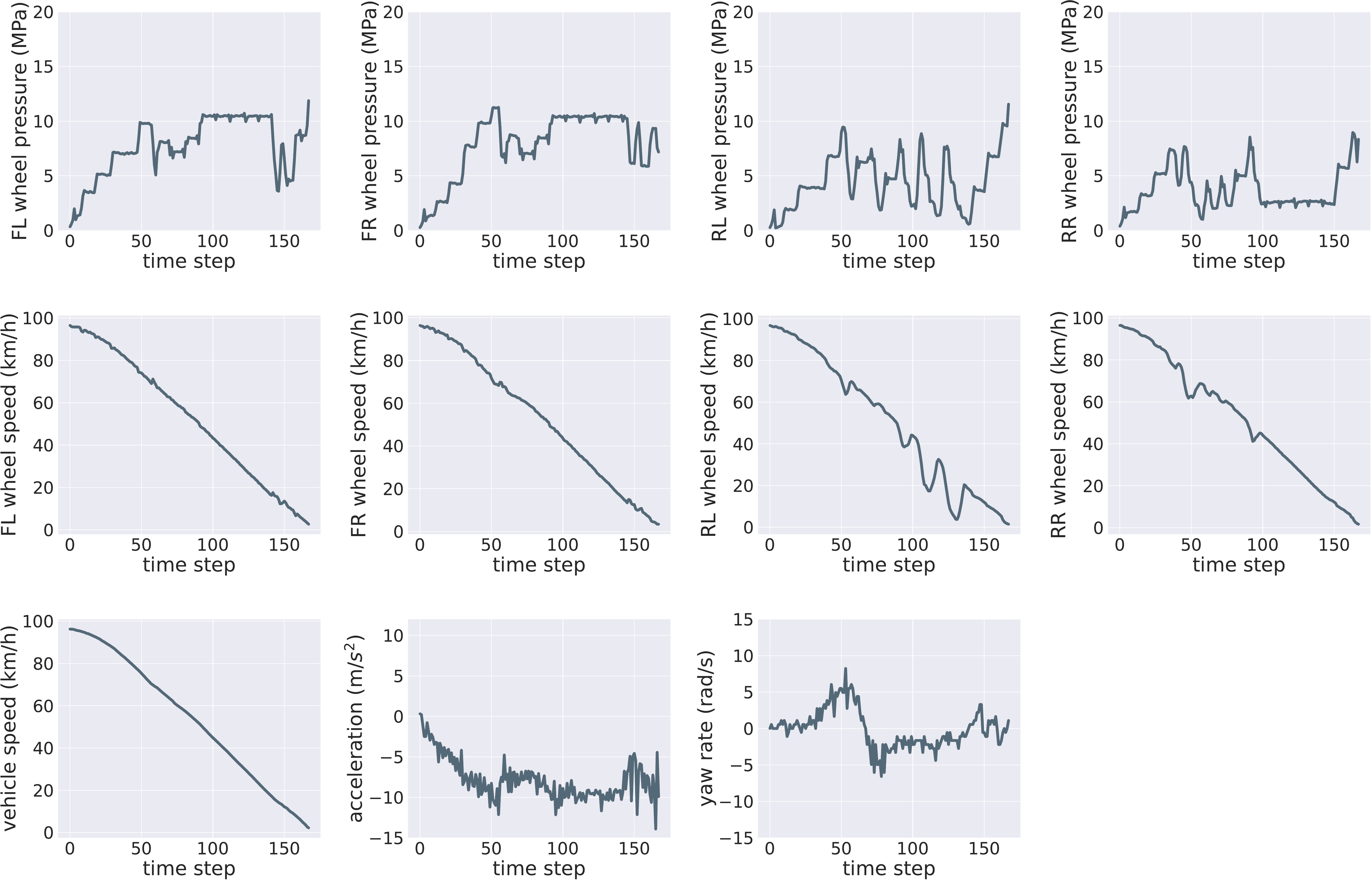}
    \caption{The braking process of the vehicle with our controller on a high-adhesion straight.}
\end{figure*}

\begin{figure*}[h!]
    \centering
    \includegraphics[width=0.88\linewidth]{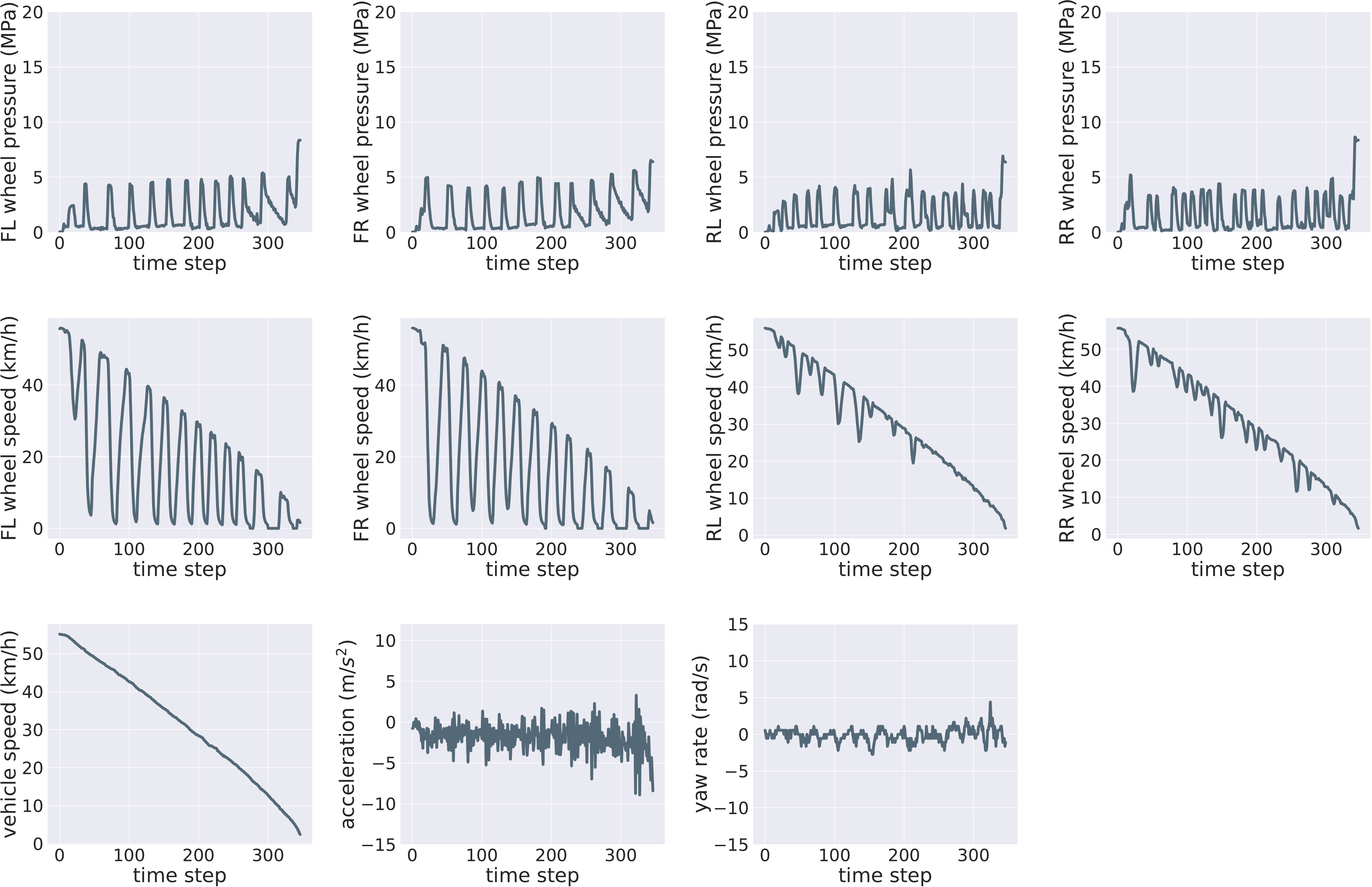}
    \caption{The braking process of the vehicle with our controller on a low-adhesion straight.}
\end{figure*}

\begin{figure*}[h!]
    \centering
    \includegraphics[width=0.88\linewidth]{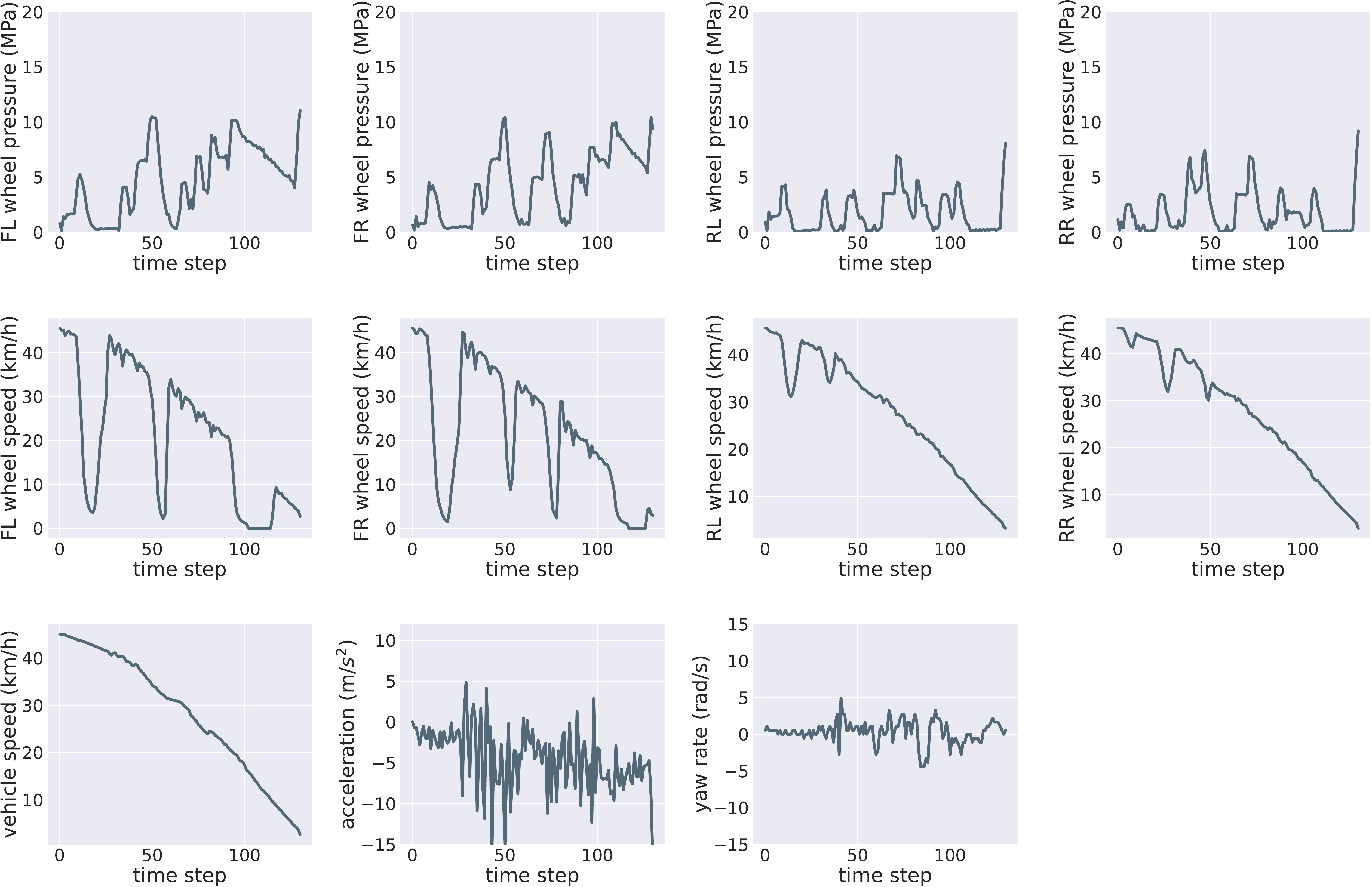}
    \caption{The braking process of the vehicle with our controller on a high-to-low straight.}
\end{figure*}

\begin{figure*}[h!]
    \centering
    \includegraphics[width=0.88\linewidth]{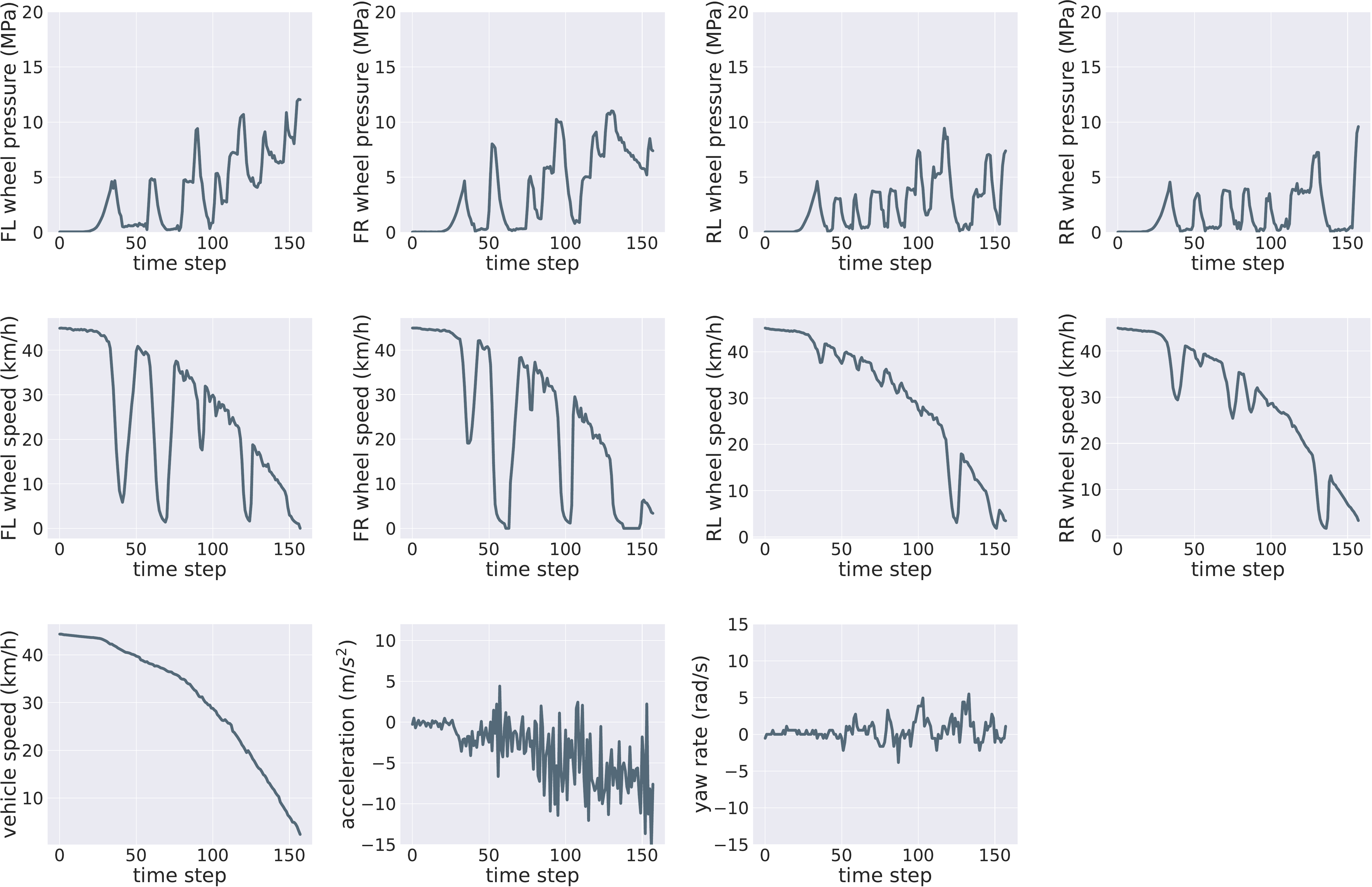}
    \caption{The braking process of the vehicle with our controller on a low-to-high straight.}
\end{figure*}

\begin{figure*}[h!]
    \centering
    \includegraphics[width=0.88\linewidth]{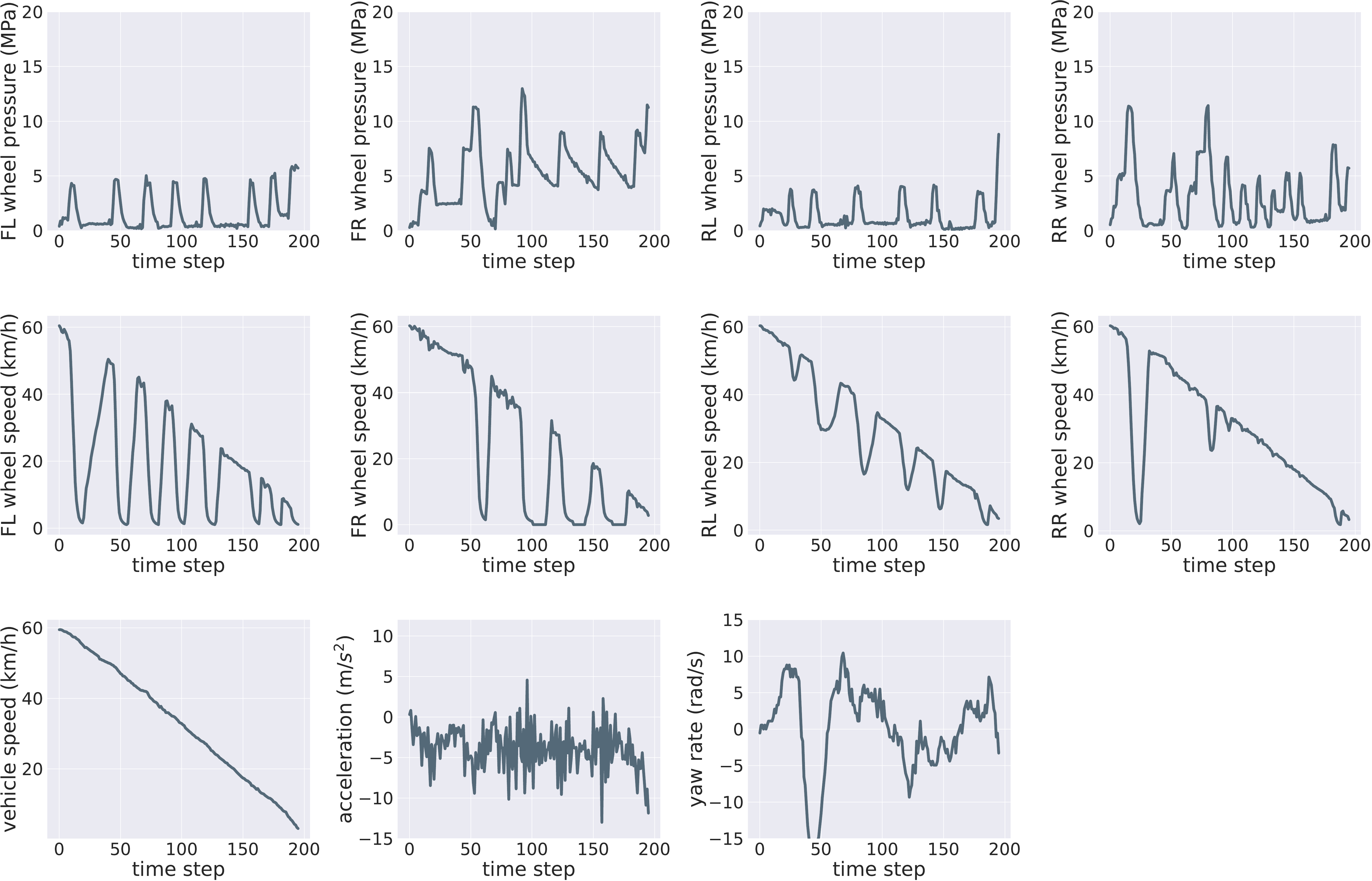}
    \caption{The braking process of the vehicle with our controller on a split-friction straight.}
\end{figure*}

\begin{figure*}[h!]
    \centering
    \includegraphics[width=0.88\linewidth]{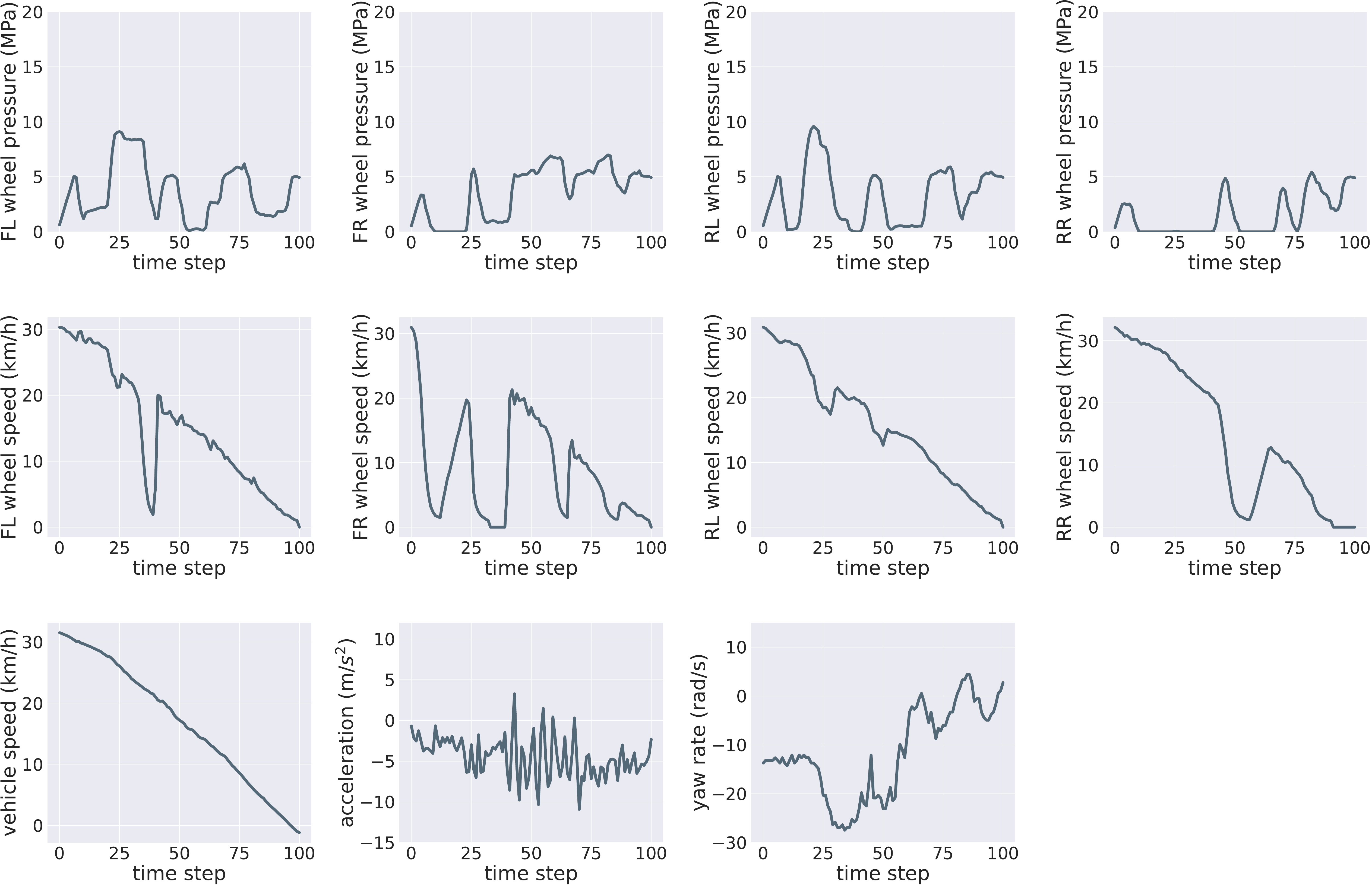}
    \caption{The braking process of the vehicle with our controller on a split-friction curve.}
\end{figure*}

\end{document}